\documentclass[10pt,twocolumn,letterpaper]{article}

\usepackage{wacv}
\usepackage{times}
\usepackage{epsfig}
\usepackage{graphicx}
\usepackage{amsmath}
\usepackage{amssymb}

\usepackage{titling}
\usepackage{subcaption}
\usepackage{algpseudocode}
\usepackage[linesnumbered,ruled]{algorithm2e}
\usepackage{booktabs}
\usepackage{nccmath}
\usepackage{array}
\usepackage{enumitem}
\usepackage[T1]{fontenc}  
\usepackage[belowskip=-15pt,aboveskip=10pt]{caption}
\newcolumntype{P}[1]{>{\centering\arraybackslash}p{#1}}

\def\wacvPaperID{497} 

\wacvfinalcopy 

\ifwacvfinal
\def\assignedStartPage{1} 
\fi

\ifwacvfinal
\else
\fi

\begin{document}

\title{Two-hand Global 3D Pose Estimation using Monocular RGB}

\author{Fanqing Lin \qquad Connor Wilhelm \qquad Tony Martinez \\Brigham Young University, Provo UT 84602, USA} 
%
%
\date{}
\maketitle

\begin{abstract}
We tackle the challenging task of estimating global 3D joint locations for both hands via only monocular RGB input images. We propose a novel multi-stage convolutional neural network based pipeline that accurately segments and locates the hands despite occlusion between two hands and complex background noise and estimates the 2D and 3D canonical joint locations without any depth information. Global joint locations with respect to the camera origin are computed using the hand pose estimations and the actual length of the key bone with a novel projection algorithm. To train the CNNs for this new task, we introduce a large-scale synthetic 3D hand pose dataset. We demonstrate that our system outperforms previous works on 3D canonical hand pose estimation benchmark datasets with RGB-only information. Additionally, we present the first work that achieves accurate global 3D hand tracking on both hands using RGB-only inputs and provide extensive quantitative and qualitative evaluation.
\end{abstract}

\section{Introduction}
\indent As the primary operating tool for human activities, the hands play a significant role in applications such as gesture control, action recognition, human-computer interaction and VR/AR. As the field of computer vision advances, commercial systems \cite{leapmotion,hololens2,htcvive,oculus} are shifting from marker/glove-based methods to vision-based hand tracking and pose estimation. However, accurate hand pose estimation from camera inputs remains challenging due to the possible heavy occlusion from the hand itself, the other hand or objects, complex background noise and the large pose space.\\
\indent Most contemporary vision-based markerless works tackling the task of 3D hand pose estimation rely on depth information, requiring either multi-view setup or depth cameras. However, such hardware requirements add severe limitations to the possible applications by significantly increasing the setup overhead and cost. Depth cameras also only work in indoor scenes and have relatively high-power consumption. To circumvent this problem, some recent approaches tackle 3D canonical one-hand pose estimation using deep CNNs with only RGB-based inputs and show good results.\\
\begin{figure}
  \begin{subfigure}[b]{\linewidth}
  \centering
  \includegraphics[width=0.9\linewidth]{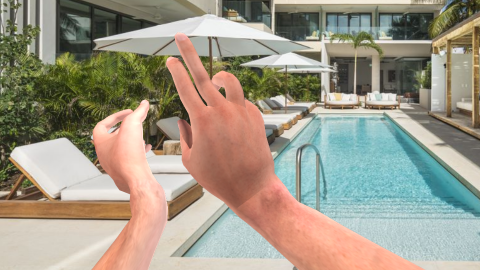}
  \end{subfigure}
  \begin{subfigure}[b]{\linewidth}
  \centering
  \vspace{0.1cm}
  \includegraphics[width=0.9\linewidth]{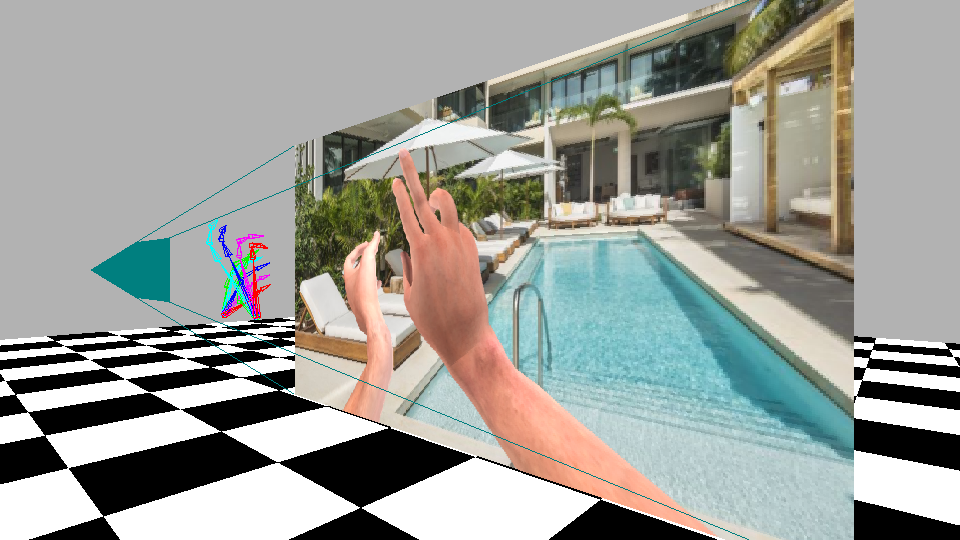}
  \end{subfigure}
  \caption{We present an approach to estimate the global 3D hand poses for both hands from a single RGB image, a new task that is particularly challenging not only due to complex background noise and various types of occlusion, but also the lack of depth information for estimating the distances. Given a RGB image (top), we show the side view of our estimation of global 3D hand poses (bottom).}
  \label{fig:intro_img}
\end{figure}
\indent In this paper, we present the first algorithm that simultaneously estimates the 3D global joint locations of both hands with respect to the camera origin using monocular RGB inputs (Fig. \ref{fig:intro_img}), which is an essential step towards the next generation gesture control and pose recognition systems. Our pipeline consists of 4 major components: (1) hand segmentation and detection, (2) 2D hand pose estimation, (3) 3D canonical hand pose estimation and (4) 3D global hand pose estimation.\\
\indent The first challenge of training our pipeline is the lack of annotated data from existing benchmark datasets. Real-world markerless 3D hand tracking data collected using multiple cameras or RGB-D camera setup inevitably has tracking error. Manual annotation is time-consuming and infeasible for large-scale data collection. Consequently, many recent benchmark datasets provide synthetic data with perfect ground truth annotation of the joint locations. However, synthetic images have different statistical distributions than real-world images and knowledge learned on synthetic data does not always transfer to the real-world domain, since CNNs are sensitive to textural information. \\
\indent Since it is currently infeasible to collect real-world two-hand pose data with accurate 2D and 3D joint annotations at a large-scale with sufficient variety, we create a novel high-quality synthetic 3D hand dataset suitable for training and evaluating the networks and focus on the challenging task of two-hand global 3D pose estimation using RGB.\\
\indent In summary, our main contributions include:
\begin{itemize}[noitemsep]
  \item The first system capable of estimating 3D global joint locations for both hands using monocular RGB inputs. We introduce a viewpoint-invariant global projection algorithm capable of perfectly reconstructing the absolute 3D joint locations and the evaluation protocol for the new task.
  \item A novel egocentric RGB-D two-part (static + dynamic) synthetic dataset for the task of two-hand 3D global pose estimation, which introduces unique challenges and also benefits researches in hand segmentation and detection, 2D and 3D canonical pose estimation. 
  \item Extensive evaluation on both two-hand 3D global and single-hand 3D canonical hand pose estimation on 4 target datasets. Our networks outperform the current state-of-the-art canonical methods with less information (only RGB) and additionally achieve promising results for global pose estimation.
\end{itemize}
\section{Related Work}
\indent Hand pose estimation is a long-standing research area due to its wide range of applications. Compared to the popular task of body pose estimation, vision-based 3D hand pose estimation has more complex articulation, heavier occlusion and more restricted availability of data. We first review the most relevant previous methods that utilize depth information, then shift our emphasis to approaches that use RGB-only data.\\
\noindent\textbf{Depth-based methods}. Oberweger \etal \cite{Oberweger} and Zhou \etal \cite{Zhou}  introduced CNN architectures that regressed 3D joint locations from depth images directly. Ge \etal \cite{Ge} proposed to project the depth image onto three orthogonal planes and fuse the corresponding 2D joint locations for the final 3D joint locations. Cai \etal \cite{Cai} and Iqbal \etal \cite{Iqbal} proposed models capable of training on RGB-D data and evaluating on RGB inputs.\\
\noindent\textbf{Multiple-camera methods}. Many methods use multiple RGB cameras to gain additional information from different viewpoints that can help with resolving the depth ambiguity and heavy occlusion. Wang \etal \cite{Wang} used 2 cameras and estimated 3D hand pose by matching with instances in a hand database. Oikonomidis \etal \cite{Oikonomidis} demonstrated the tracking of both the hand and an interacting object in 3D with 8 surrounding fixed cameras. Sridhar \etal \cite{Scridhar,Scridhar2} estimated the hand pose by using generative approach on inputs of multiple RGB cameras and a depth sensor. Zhang \etal \cite{Zhang} introduced 3D hand pose estimation using matching algorithm on inputs from stereo cameras.\\
\noindent\textbf{Single-camera methods}. Due to the significantly higher setup overhead and costs introduced by depth sensors and multiple calibrated cameras, some methods use a single RGB image to estimate the 3D hand poses. Zimmermann and Brox \cite{Zimmermann} proposed a CNN-based pipeline that estimates the 2D joint locations and lifts the 2D heatmaps to 3D canonical joint locations. Mueller \etal \cite{Mueller} introduced a model that estimates both 2D and 3D canonical joint locations with kinematic skeleton fitting to better address physical constraints and temporal smoothness. Spurr \etal \cite{Spurr} and Yang \etal \cite{Yang} proposed to use variational autoencoders to learn a latent space for hand poses, which are capable of estimating 3D hand poses from RGB inputs. Some methods \cite{Baek,Xiang,Ge2} estimated the low-dimensional parameters for a 3D deformable hand model \cite{Romero} to fit the RGB inputs in order to retrieve the 3D canonical hand poses. \\
\noindent\textbf{Two-hand methods}. It is more natural yet difficult to estimate poses for two interacting hands due to inter-hand occlusion. Tzionas \etal \cite{Tzionas}, Taylor \etal \cite{Taylor} and Mueller \etal \cite{Mueller3} achieved promising results using energy optimization to fit parametric hand models using depth data. Note that 3D pose estimation using RGB data is much more challenging due to the lack of depth information and the additional noise from images in the wild. To the best of our knowledge, our work is the first to address two-hand global pose estimation using RGB data from a single camera.
\section{Datasets for Hand Pose Estimation}
\indent For depth-based hand pose estimation, \cite{Tompson,Tang,Sun,Yuan,Garcia} presented large-scale datasets consisting of real depth images with estimated ground-truth joint locations. For RGB-D hand pose estimation, due to the need to manually annotate joint locations, small-scale datasets \cite{Mueller2,Sridhar3,Zhang} with limited variation were presented with real RGB and depth data. For RGB-based hand pose estimation, \cite{Simon, Zimmermann2} performed extensive manual annotation and provided a decent amount of labeled real-world instances. Note that accurately annotated hand data with sufficient variation is necessary for learning-based approaches, and datasets with real-world RGB/depth data can only provide estimation of the joint locations as the ground truth and some severely lacks sufficient variety. Consequently, synthetic datasets \cite{Mueller,Mueller2,Zimmermann} with large amount of color, depth images and perfect annotation are introduced for advancing research in the field. It is worth mentioning that existing RGB-based datasets and methods are designed to estimate the 3D hand poses in a canonical (localized) frame for a single hand. Therefore, it is necessary to generate a new dataset suitable for the task of RGB-based two-hand global pose estimation.
\subsection{Ego3DHands Dataset}
\begin{figure}[t]
  \centering
  \begin{subfigure}[b]{0.48\linewidth}
  \includegraphics[width=\linewidth]{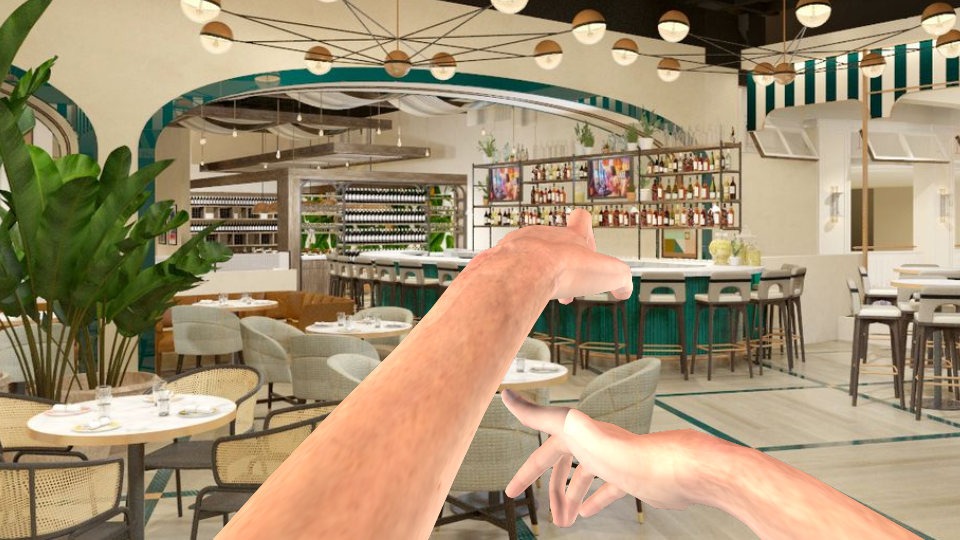}
  \end{subfigure}
  \begin{subfigure}[b]{0.48\linewidth}
  \includegraphics[width=\linewidth]{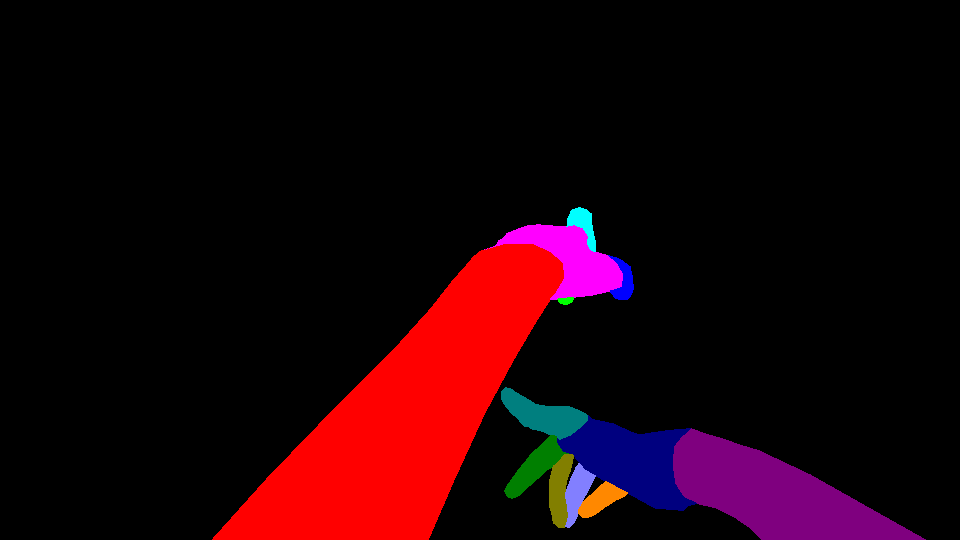}
  \end{subfigure}
  \caption{Our dataset provides a total of 14 segmentation masks (right) for the fingers, palm and arm along with the 2D and 3D joint annotations.}
  \label{fig:dataset_img}
\end{figure}

\indent We introduce the first dataset for the task of two-hand global 3D pose estimation from an egocentric view. Following \cite{Zimmermann}, the dataset is generated using rendering from \textit{Blender}\footnote{www.blender.org}, which enables us to obtain the segmentation masks of hand parts as well as the annotated 2D and 3D joint locations (infeasible to obtain on real hand data at large-scale with variety). We utilize a single character from \textit{Mixamo}\footnote{www.mixamo.com} to keep the bone ratios of the hands consistent for global 3D pose reconstruction. The dataset includes two versions for static and dynamic hand pose estimation respectively. Despite the domain gap between synthetic and real-world data, this dataset enables training for learning-based approaches and quantitative analysis for a new task. \\
\noindent\textbf{Data Representation}. As illustrated in Fig. \ref{fig:dataset_img}, the dataset provides 7 segmentation masks for each hand. The 2D joint locations are normalized values ranging from (0, 0) at the top left to (1, 1) at the bottom right. The global 3D coordinates are represented in the camera space. We scale the 3D coordinates so that the bone length from the wrist to the middle metacarpophalangeal (mMCP) is 10 cm. Each hand consists of 21 joints: wrist and 4 joints for each finger. The Depth map is also provided but not used in this work.\\
\noindent\textbf{Static Ego3DHands}. To capture the images for static hand poses, we set the camera to be between the eyes of the character facing forward. We keep the hands inside a fix-sized bounding box in front of the character so the targets stay in sight for estimation. Each hand has a 10\% drop rate for single-hand scenarios. The rotational angles for arm and hand joints are randomized within reasonable rotational ranges to obtain vast variety in the pose space. We include 4 light sources with slightly randomized color, brightness and position for illumination. Additionally, for the background of the hand pose images, we selected 100 unique scene topics and collected 20,000 images from online sources, on which we further applied random color augmentation and horizontal flips. We create 50,000 instances for the training set and 5,000 instances for the test set.
\\
\noindent\textbf{Dynamic Ego3DHands}. For global dynamic two-hand 3D tracking, we introduce an additional dataset with 100 sequences for the training set, and 10 sequences for the test set. Each sequence consists of 500 frames where we randomize independent motion for both hands. For background sequences, we selected 110 short videos with variety from Pexels\footnote{www.pexels.com}, so each hand pose sequence has a unique corresponding background sequence. This dataset enables researchers to explore methods for 3D global hand pose estimation that utilize temporal consistency. We report our baseline results in Section \ref{sec:experiments} for future comparison.
\begin{figure*}[t!]
  \centering
  \begin{subfigure}[b]{1.0\linewidth}
    \includegraphics[width=\linewidth]{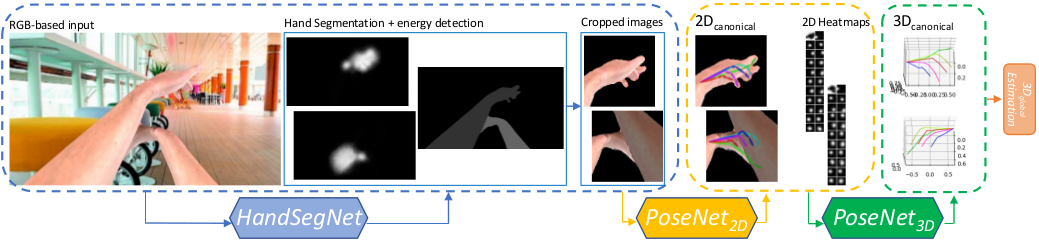}
  \end{subfigure}
  \caption{Overview of proposed pipeline for two-hand global pose estimation from monocular RGB. Given a RGB image, we segment and detect the hands, crop and process the hand images for 2D canonical pose estimation despite inter-hand occlusion, and use the 2D heatmaps to estimate 3D canonical poses. For the final step, We introduce a novel algorithm for computing the 3D global hand poses using the 2D and 3D canonical estimation as well as the actual bone lengths.}
  \label{fig:method_img}
\end{figure*}
\section{Method}
\indent In this paper, we present the first algorithm capable of estimating the global 3D poses of both hands from a monocular RGB image. The overall system is demonstrated in Fig. \ref{fig:method_img}. Given a single RGB image as input, we use \textit{HandSegNet} to simultaneously obtain the segmentation masks and the heatmap energy of both hands. The hand heatmap energy indicates the approximate locations of the hands despite occlusion, which are used to detect and provide a cropped image for each hand. The cropped RGB hand images are then processed using the corresponding segmentation masks for the next stage. To estimate the 2D joint locations, we present \textit{$PoseNet_{2D}$} that estimates and refines the 2D heatmaps of the joints in multiple stages. To lift the 2D heatmaps to a 3D pose estimation, we present \textit{$PoseNet_{3D}$} that takes the heatmaps as input and regresses the 3D canonical joint locations which we define in detail in Section \ref{sec:3d_can}. Finally, we present a novel algorithm that accurately estimates the 3D global hand joint locations in the spherical coordinate system using the obtained 2D and 3D canonical information, the actual length of key bone and the camera intrinsics. Our method can theoretically be applied to estimate the global location and pose of any object given the aforementioned information.
\subsection{Two-hand Segmentation and Detection}
\indent Unlike existing methods that perform pose estimation on a single cropped hand, we need to first distinguish between left and right hand by estimating the individual hand locations. For the task of hand segmentation and detection, we use a deep convolutional neural network trained to predict both the segmentation masks and the location of hands in form of heatmap energy. We show in Section \ref{sec:2D} that the accurate segmentation of hands is necessary information for 2D hand pose estimation in the presence of occlusion from the other hand.\\
\indent For the architecture of \textit{HandSegNet}, we use a residual network for the task of semantic segmentation and hand detection. It consists of 4 downsampling and 4 upsampling layers comprised of 16 residual blocks. For the output layer, we have 3 channels for the task of segmentation (2 objects and background) and 2 additional channels for estimating the heatmap energy of the left and right hand. The heatmap energy is capable of providing high activation at locations of partially or even completely occluded hands. To generate the cropped bounding boxes, we apply Otsu Thresholding on the hand energy for selecting the high activation area. In the case of very low activation, we classify the corresponding hand(s) as being absent and drop the absent hand(s) in the subsequent stages.
\subsection{2D Canonical Hand Pose Estimation}\label{sec:2D}
\begin{figure}[t]
  \centering
  \begin{subfigure}[t]{0.19\linewidth}
    \includegraphics[width=\linewidth]{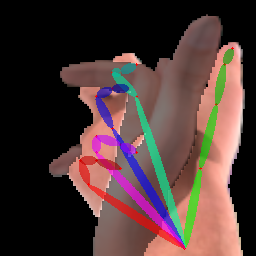}
  \end{subfigure}
  \begin{subfigure}[t]{0.19\linewidth}
    \includegraphics[width=\linewidth]{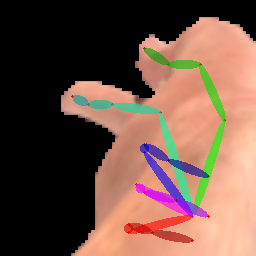}
  \end{subfigure}
  \begin{subfigure}[t]{0.19\linewidth}
    \includegraphics[width=\linewidth]{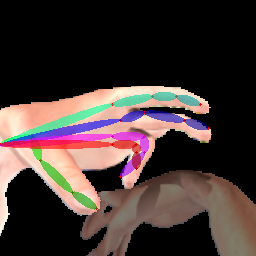}
  \end{subfigure}
  \begin{subfigure}[t]{0.19\linewidth}
    \includegraphics[width=\linewidth]{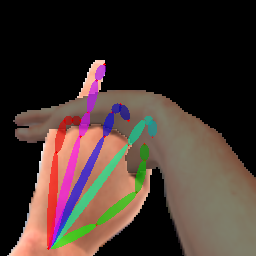}
  \end{subfigure}
  \begin{subfigure}[t]{0.19\linewidth}
    \includegraphics[width=\linewidth]{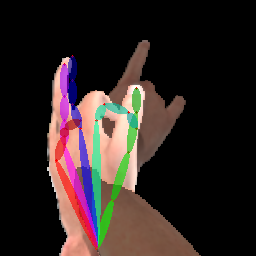}
  \end{subfigure}\\
  \begin{subfigure}[t]{0.19\linewidth}
    \includegraphics[width=\linewidth]{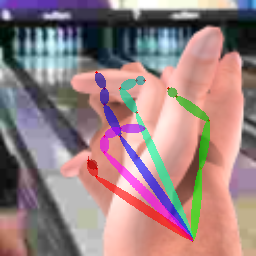}
  \end{subfigure}
  \begin{subfigure}[t]{0.19\linewidth}
    \includegraphics[width=\linewidth]{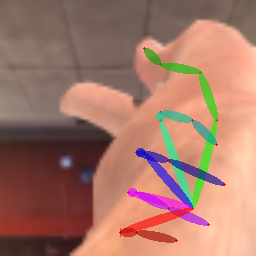}
  \end{subfigure}
  \begin{subfigure}[t]{0.19\linewidth}
    \includegraphics[width=\linewidth]{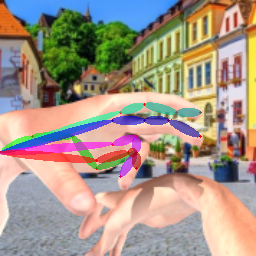}
  \end{subfigure}
  \begin{subfigure}[t]{0.19\linewidth}
    \includegraphics[width=\linewidth]{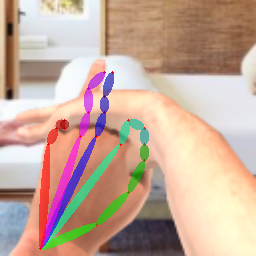}
  \end{subfigure}
  \begin{subfigure}[t]{0.19\linewidth}
    \includegraphics[width=\linewidth]{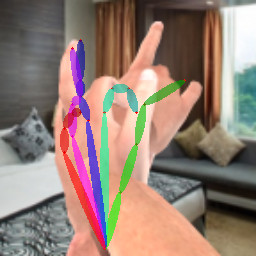}
  \end{subfigure}
  \caption{We show that 2D canonical hand pose estimation with segmentation information (top) better resolves inter-hand occlusion from the secondary hand.}
  \label{fig:seg_2d_demo_img}
\end{figure}

\indent The goal of \textit{$PoseNet_{2D}$} is to estimate the 2D joint locations given a cropped hand image. We use a variant of Convolutional Pose Machines (CPM) \cite{Wei} as our base model, with batch normalization layers inserted after the convolutional layers for better adaptation to the vast RGB image space. The 2D joint locations are represented as heatmaps and CPM refines the output heatmaps in progressive stages. Since the left and right hand have different articulation, we horizontally flip the cropped images of the right hand so the learned articulation remains consistent for the model. We resize the cropped input hand images from \textit{HandSegNet} to 256x256. The output of the CPM consists of 21 heatmaps with size of 32x32. We generate stronger heatmap energy for closer joints so that depth information is encoded into the 2D heatmaps. We show that this heatmap generation technique (we refer to as \textit{z-heatmaps}) improves accuracy for 3D canonical hand pose estimation in Section \ref{sec:experiments}.\\
\indent Previous work \cite{Mueller} showed good performance on RGB-based 2D hand pose estimation with occlusion introduced by an interacting object. However, the method fails when the background has similar appearance as the hand. With the presence of both hands, accurate 2D hand pose estimation becomes more challenging due to the similar-object occlusion introduced by the secondary hand. We show in Fig. \ref{fig:seg_2d_demo_img} that we successfully address this issue by providing the segmentation information necessary for distinguishing the left and right hand. By using the segmentation masks of the two hands, we simplify the input image space by removing the background noise; additionally, we differentiate the color space between two hands by reducing the brightness of the secondary hand by a factor of 0.5. As a result, it is very important for \textit{HandSegNet} to produce accurate segmentation masks for the two hands. The output heatmaps are used as inputs for \textit{$PoseNet_{3D}$} and we retrieve the resulting 2D global joint locations using the bounding box coordinates from \textit{HandSegNet}. For a set of 2D global joint locations, we use $\textbf{p}_{j} = (\rho^{r}_{j}, \rho^{c}_{j})$, where $\rho^{r}$ and $\rho^{c}$ represent the corresponding row and column position of the $j^{th}$ joint in form of percentages.
\subsection{3D Canonical Hand Joints Regression}\label{sec:3d_can}
\indent The 3D canonical frame for a single hand is defined such that the middle metacarpophalangeal (mMCP) joint is at the origin and the distance between wrist and mMCP is 1 \cite{Mueller}. The defined canonical frame requires the target hand to be in the middle of the cropped image so the z-axis aligns with the camera direction. Therefore, we generate the 3D canonical joint annotation for training by rotating the original global 3D joint locations of both hands to the center of the image; zero-centering on the mMCP and normalization is applied afterwards. Thus, for a set of annotated canonical 3D Cartesian coordinates represented as $\textbf{w}_{j} = (\textit{x}_{j}, \textit{y}_{j}, \textit{z}_{j})$, \\
\begin{equation}\label{eq:3d_canonical}
\begin{gathered}
\textbf{w}^{center} = \mathcal{R}\cdot\textbf{w}^{glob}\\
\textbf{w}^{can} = (\textbf{w}^{center} - \textbf{w}^{center}_{mmcp}) / \textbf{\textit{d}}
\end{gathered}
\end{equation}
where $\mathcal{R}$ is the 3D rotational matrix for centering $\textbf{w}^{can}_{mmcp}$ and \textbf{\textit{d}} is the distance between the wrist joint and the mMCP. As a result, our 3D canonical hand poses are consistent with the visual representation of the hands, which is necessary for estimating the global joint locations and will be explained in Section \ref{sec:3D_glob}.\\
\indent For the architecture of \textit{$PoseNet_{3D}$}, we use a small residual network comprised of 8 residual blocks with 2 fully connected layers before the output layer. The input heatmaps are upscaled by a factor of 2 to the size of 64x64 for better performance. The model estimates the root-relative 3D coordinates of 21 joints for each hand.\\
\indent To enforce physical constraints and encourage 2D pose consistency between $\textit{w}^{can}$ and \textbf{p}, we employ the following loss function for training \textit{$PoseNet_{3D}$}, \\
\begin{equation}\label{eq:3d_loss}
\begin{gathered}
\mathcal{L}_{3d} = \mathcal{L}_{j} + \mathcal{L}_{bone} + \mathcal{L}_{proj}\\
\end{gathered}
\end{equation}
where $\mathcal{L}_{j}$ is the Mean Squared Error (MSE) loss for joint regression. In addition, we introduce $\mathcal{L}_{bone}$ that indicates the MSE between the ground truth and the predicted bone lengths. $\mathcal{L}_{proj}$ indicates the MSE between the $(x, y)$ component of $\textbf{w}^{can}$ and \textbf{p} projected into the same canonical frame. The overall $\mathcal{L}_{3d}$ aims to produce physically plausible 3D canonical hand poses consistent with the 2D poses.\\
\indent For the task of 3D canonical hand pose estimation only (Section \ref{sec:experiments_can}), we expect \textit{$PoseNet_{3D}$} to refine the potentially inaccurate 2D pose estimations and drop $\mathcal{L}_{proj}$. However, for the task of 3D global hand pose estimation in the next stage, we replace \textbf{p} from \textit{$PoseNet_{2D}$} with the $(x, y)$ component of $\textbf{w}^{can}$ projected back into the pixel space since the consistency between 2D and 3D poses is important for our projection algorithm (Section \ref{sec:3D_glob}).
\begin{figure}[t]
  \centering
  \begin{subfigure}[b]{0.9\linewidth}
    \includegraphics[width=\linewidth]{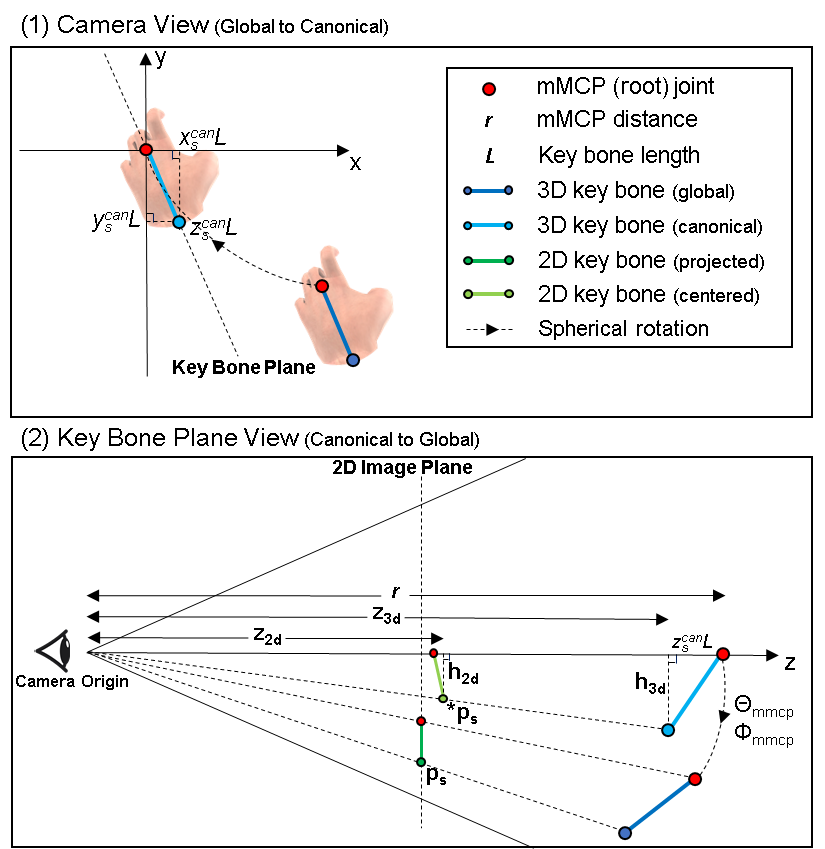}
  \end{subfigure}
  \caption{The camera view (top) shows how we treat the hand as if it is in the center of the image and generate the corresponding 3D canonical pose using Eq. (\ref{eq:3d_canonical}) for training. The bone plane view (bottom) illustrates how we compute the absolute distance \textbf{\textit{r}} of the centered root joint and subsequently return the pose to its original global 3D position by rotation in spherical coordinate system.}
  \label{fig:radius_img}
\vspace{-1mm}
\end{figure}
\begin{figure*}[t]
  \centering
  \begin{subfigure}[t]{0.32\linewidth}
    \includegraphics[width=\linewidth]{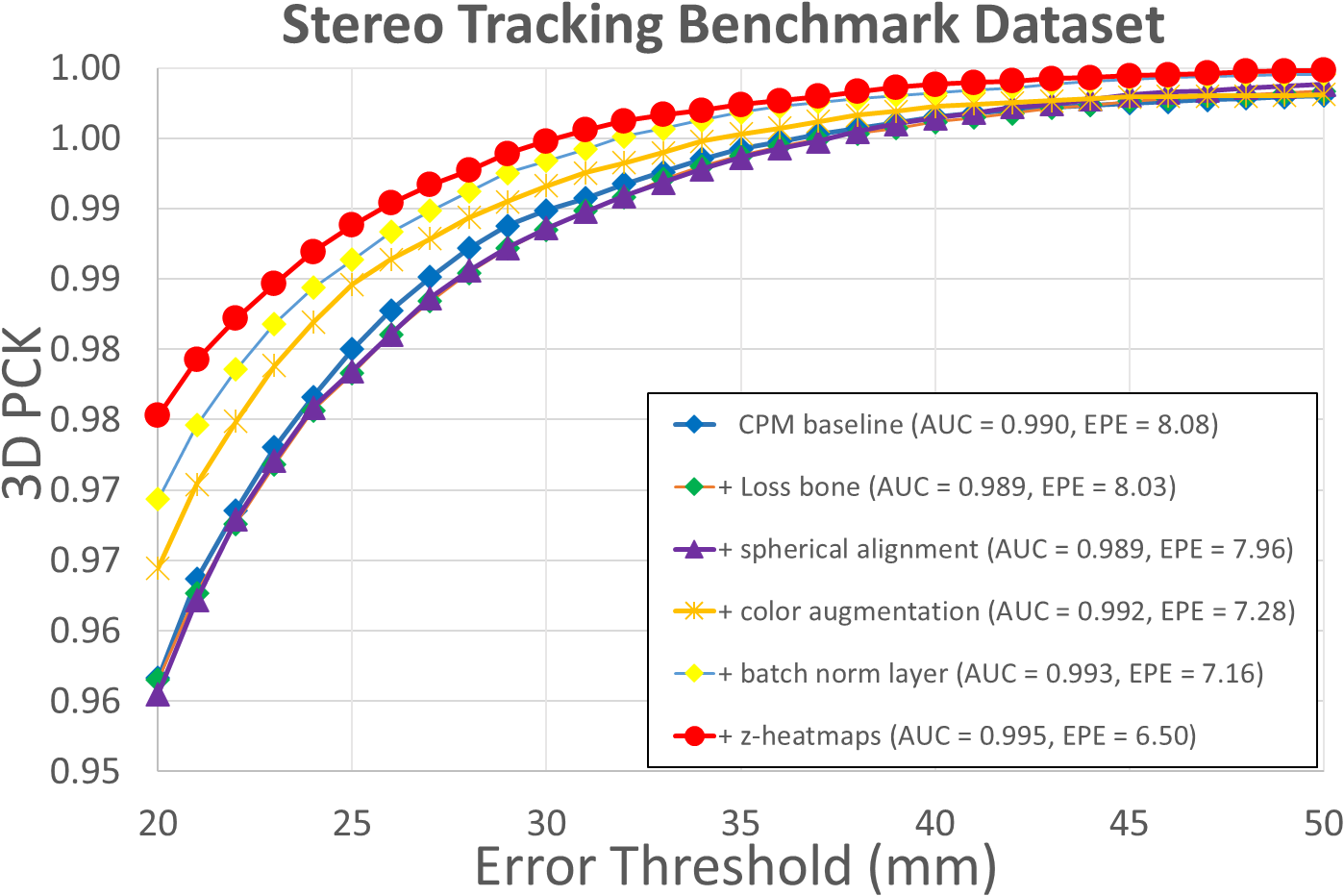}
    \caption{Self-comparisons.}\label{fig:stereo_eval1}
  \end{subfigure}
  \begin{subfigure}[t]{0.32\linewidth}
    \includegraphics[width=\linewidth]{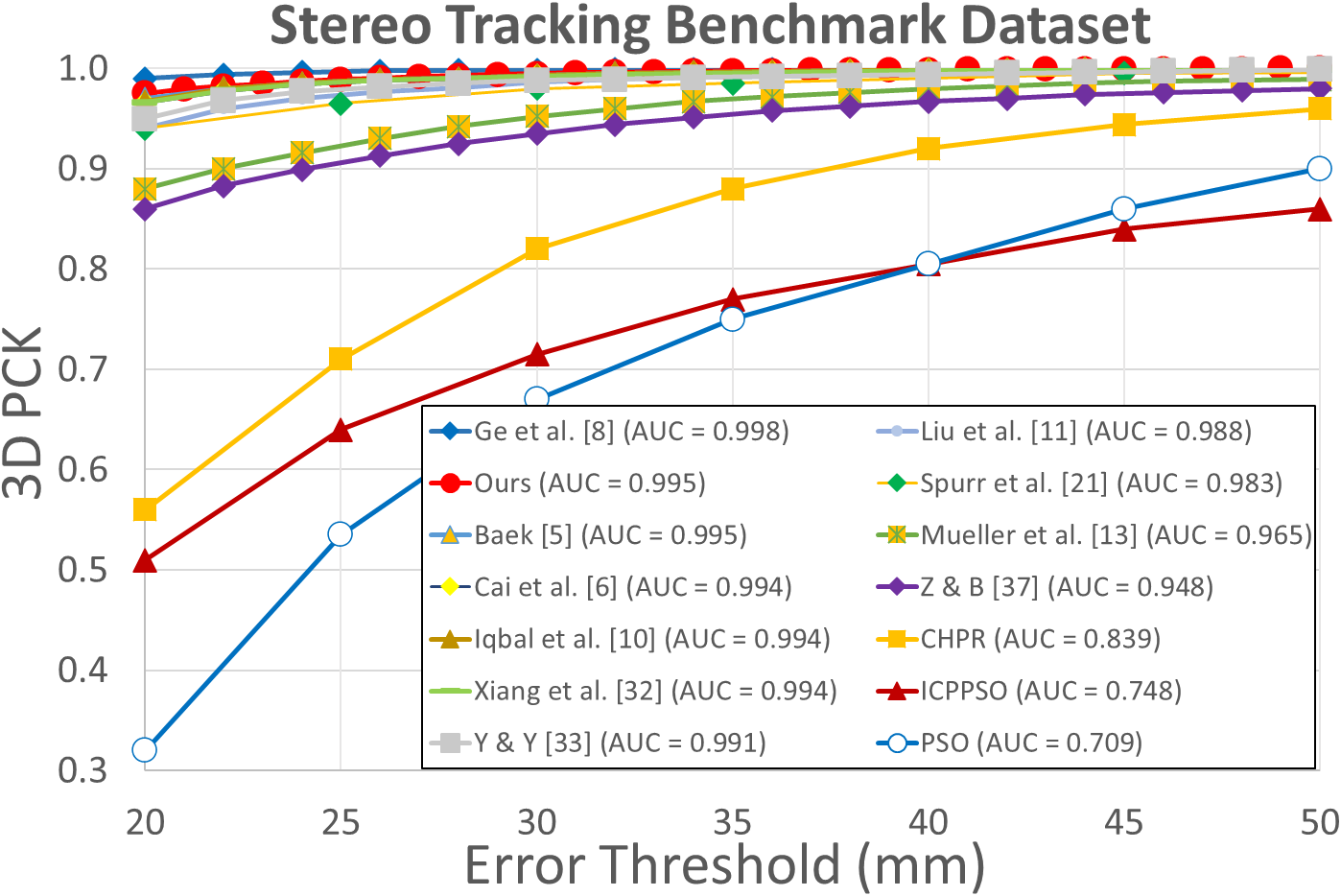}
    \caption{Comparison with the state-of-the-art.}\label{fig:stereo_eval2}
  \end{subfigure}
  \begin{subfigure}[t]{0.32\linewidth}
    \includegraphics[width=\linewidth]{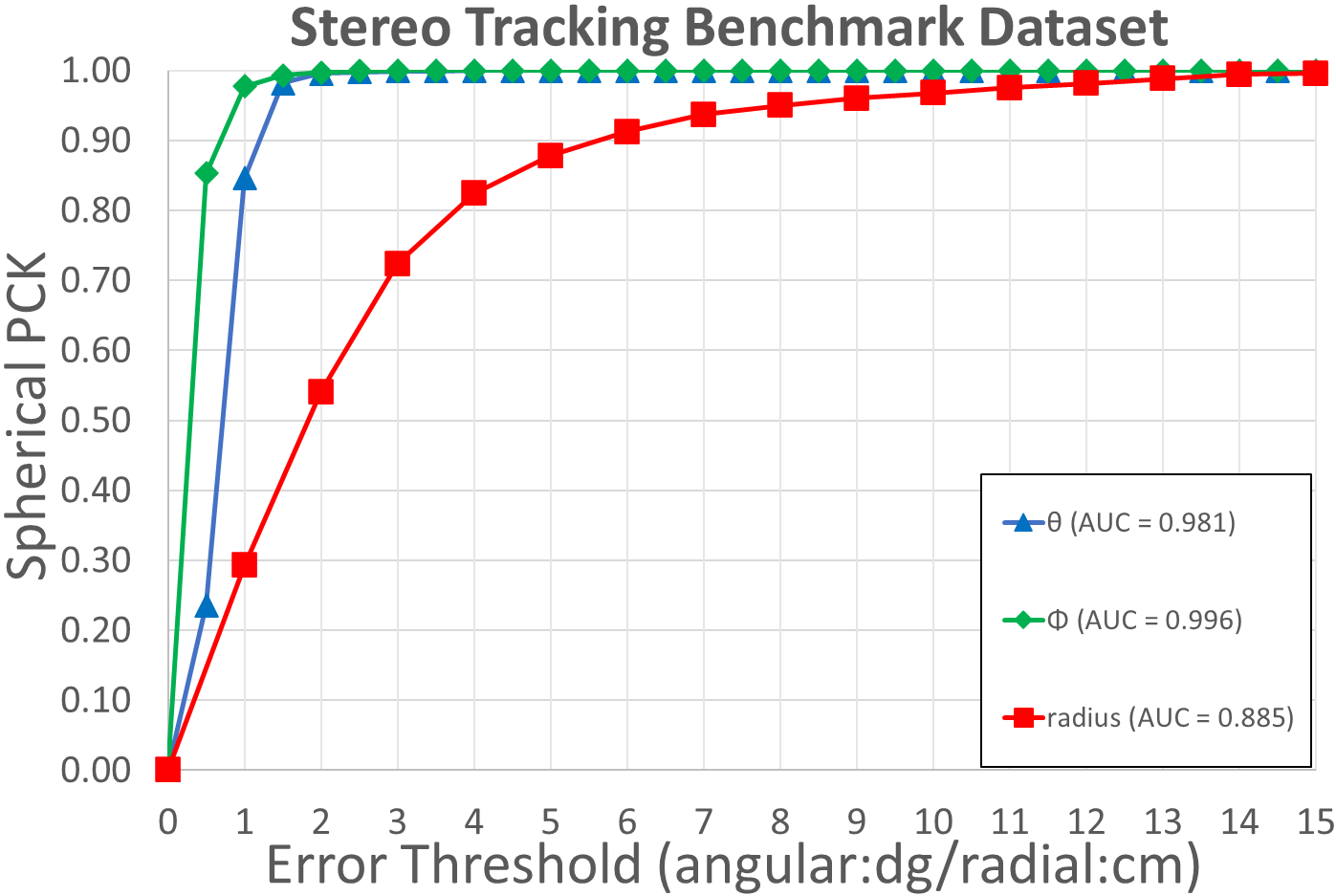}
    \caption{Spherical PCK for global estimation.}\label{fig:stereo_eval3}
  \end{subfigure}
  \vspace{0.2cm}
  \caption{Self-comparisons (left) and comparison with the state-of-the-art (middle) for 3D canonical hand pose estimation on the STB dataset. + indicates that the feature is applied incrementally. Spherical PCK (right) without alignment with the ground truth root joint is also reported.}
  \label{fig:plots_eval}
\end{figure*}
\subsection{3D Global Hand Pose Estimation}\label{sec:3D_glob}
\indent The problem of 3D global hand pose estimation in Cartesian coordinate system introduces different challenges compared to conventional 3D canonical hand pose estimation. First, the same canonical hand pose in different global positions is visually rotated, thus introducing rotational ambiguity. Second, the size of the hand correlates not to the z-value of its global Cartesian 3D position, but to its absolute distance to the camera origin, which introduces depth ambiguity. As a result, we propose a novel algorithm for global 3D hand pose estimation using the spherical coordinate system. As illustrated in Fig. \ref{fig:radius_img}, in order to transform the 3D canonical hand pose back to its original 3D global position, we first scale it by the known actual key bone length \textbf{\textit{L}} to the real-world size, then translate it by \textbf{\textit{r}} cm in the positive direction along the z-axis, and finally apply a 3D global rotation with $\theta_{mmcp}$ and $\phi_{mmcp}$ respectively in the spherical coordinate system.  To compute a set of global 3D Cartesian coordinates $\textbf{w}^{glob}$ given $\textbf{w}^{can}$ in 3D and $\textbf{p}$ in 2D, we find the absolute spherical coordinate of the mMCP $\textbf{v}_{mmcp} = (\textit{r}_{mmcp}, \theta_{mmcp}, \phi_{mmcp})$. For the rotational angles, \\
\begin{equation}\label{eq:spherical_angles}
\begin{gathered}
\theta_{mmcp} = atan(((\textbf{p}^{r}_{mmcp}\cdot\textbf{\textit{H}})-\textbf{\textit{H}}/2)/\textit{pxcm},\ \textit{foc})\\
\phi_{mmcp} = atan(((\textbf{p}^{c}_{mmcp}\cdot\textbf{\textit{W}})-\textbf{\textit{W}}/2)/\textit{pxcm},\ \textit{foc})\\
\end{gathered}
\end{equation}
where \textbf{\textit{H}} and \textbf{\textit{W}} represent the height and width of the RGB input image, \textit{pxcm} is a constant conversion factor for converting from image pixels to centimeters and \textit{foc} is the camera focal length.\\
\indent In order to compute $\textit{r}_{mmcp}$, we need to apply the Side Splitter Theorem on the right-angled similar triangles shown on the key bone plane (Fig. \ref{fig:radius_img} (2)), \\
\begin{equation}\label{eq:spherical_radius1}
\begin{gathered}
z_{3d} = z_{2d} \cdot h_{3d} / h_{2d}\\
\end{gathered}
\end{equation}
where $z_{2d}$ and $h_{2d}$ are computed by rotating $\textbf{p}_{s}$ (the secondary joint \textit{s} that forms the key bone with mMCP) with $-\theta_{mmcp}$ and $-\phi_{mmcp}$. For $h_{3d}$, the segment of key bone perpendicular to the z-axis is\\
\begin{equation}\label{eq:spherical_radius2}
\begin{gathered}
h_{3d} = \sqrt{{x^{can}_{s}}^2 + {y^{can}_{s}}^2}\cdot\textbf{\textit{L}}\\
\end{gathered}
\end{equation}
where $(x_{s}^{can}, y_{s}^{can}, z_{s}^{can})$ is the 3D canonical position for joint \textit{s}. Finally, we compute the spherical radius of the mMCP, \\
\begin{equation}\label{eq:spherical_radius3}
\begin{gathered}
\textit{r}_{mmcp} = z_{3d} - z_{s}^{can} \cdot \textbf{\textit{L}}\\
\end{gathered}
\end{equation}
with $z_{s}^{can}$ being positive if the key bone extends away from the camera.\\
\indent Since the key bone needs to have sufficient length in 2D images for accurate projection and estimation, we use mMCP as the primary joint for the key bone, and select either the wrist or the pinky MCP as the secondary joint. By selecting the longer one of the two \textquotedbl bones\textquotedbl\ as the key bone, we guarantee its validity since the two \textquotedbl bones\textquotedbl\ can never be parallel and therefore can never both point in the direction of z-axis in 2D images.\\
\indent For tracking both hands in 3D global space in video settings with temporal smoothness, we apply polynomial regression for the estimation of $\textit{r}_{mmcp}$. We use two queues (for left and right hands) to store the most recent $\textit{r}_{mmcp}$ values to estimate the current $\textit{r}_{mmcp}$.\\
\indent Our 3D global pose estimation algorithm can be applied generally to estimate the 3D global location of any objects given the 2D, 3D canonical information, actual key bone length and camera intrinsics. Unlike other methods \cite{Moon,Rogez} that attempt to estimate the approximate global poses, our algorithm is, to our knowledge, the first capable of perfectly reconstructing the global 3D poses given accurate input information. We show its effectiveness in Section \ref{sec:experiments_glob} on several hand pose datasets (both synthetic and real) with annotated 3D global joint locations and different viewpoints.
\section{Experiments}\label{sec:experiments}
\indent We first compare the performance of \textit{$PoseNet_{2D}$} and \textit{$PoseNet_{3D}$} for single-hand 3D canonical pose estimation with the current state-of-the-art methods on two popular benchmark datasets: Stereo Tracking Benchmark Dataset (STB) \cite{Zhang} and Rendered Hand Pose Dataset (RHP) \cite{Zimmermann}. For two-hand global 3D pose estimation, we evaluate our method on the test sets of both the static ($Ego3D_{s}$) and the dynamic ($Ego3D_{d}$) versions of Ego3DHands. To demonstrate the effectiveness of our global pose estimation algorithm, we show quantitative and qualitative results for global hand pose estimation on all 4 target datasets. Training details are included in the supplementary document.
\begin{figure*}[t]
 \centering
  \begin{subfigure}[t]{0.32\linewidth}
    \includegraphics[width=\linewidth]{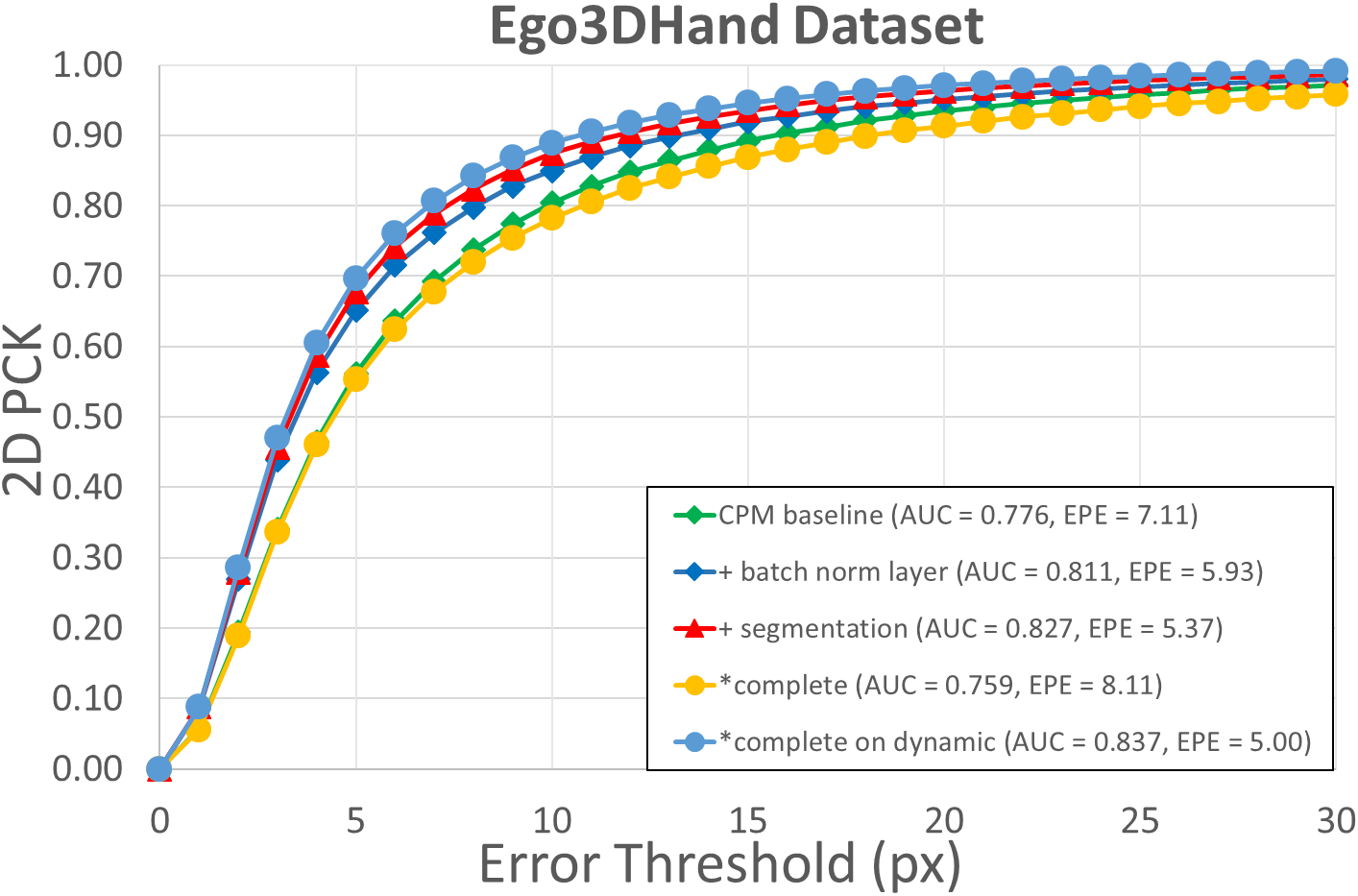}
    \caption{2D Pose Estimation.}\label{fig:ego3d_eval1}
  \end{subfigure}
  \begin{subfigure}[t]{0.32\linewidth}
    \includegraphics[width=\linewidth]{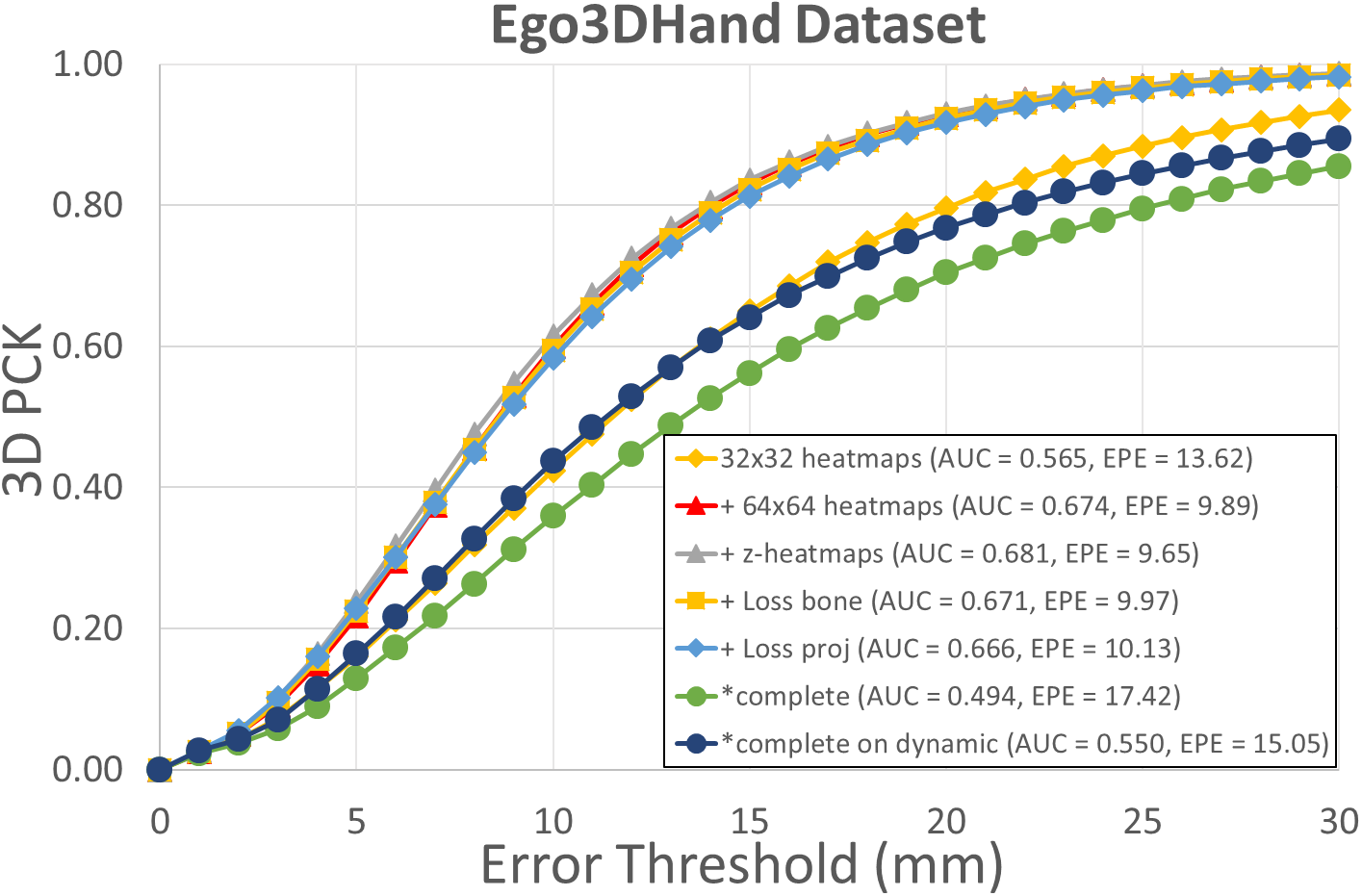}
    \caption{3D Canonical Pose Estimation.}\label{fig:ego3d_eval2}
  \end{subfigure}
  \begin{subfigure}[t]{0.32\linewidth}
    \includegraphics[width=\linewidth]{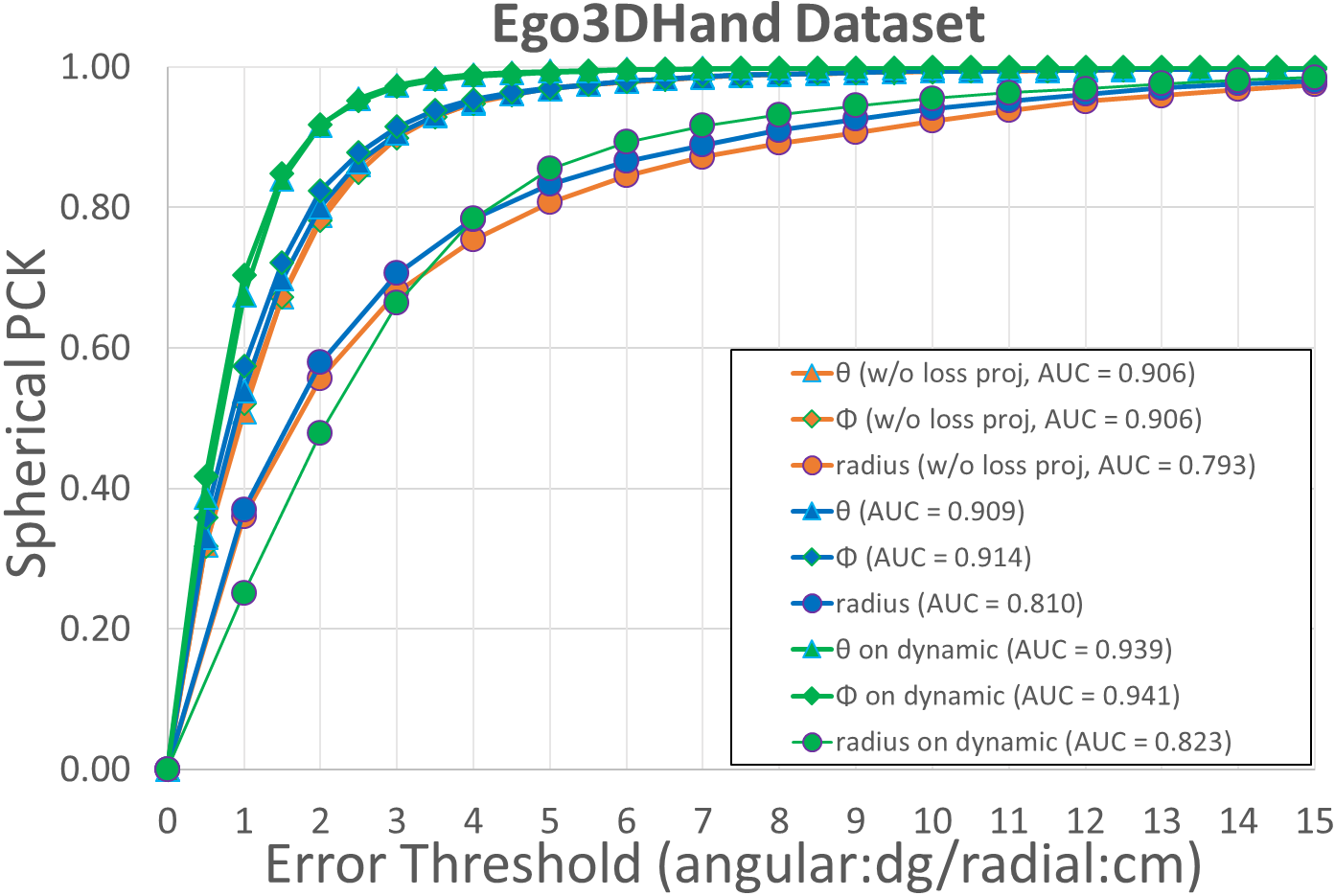}
    \caption{Spherical PCK for global estimation.}\label{fig:ego3d_eval3}
  \end{subfigure}
  \vspace{0.2cm}
  \caption{Quantitative results on the $Ego3D_{s}$ and $Ego3D_{d}$ for 2D (a), 3D canonical (b) and 3D global (c) hand pose estimation. 2D PCK is computed using image size of (270x480). Results are reported given the ground truth input for isolated studies unless marked as \textquotedbl *complete\textquotedbl \text{ in (a) and (b)}. \textquotedbl dynamic\textquotedbl \ indicates that experiments are performed on $Ego3D_{d}$.}
  \label{fig:ego3d_eval}
\end{figure*}
\subsection{Single-hand Canonical Pose Estimation}\label{sec:experiments_can}
\indent Stereo Tracking Benchmark Dataset (STB) consists of 12 sequences (1500 frames per sequence) of captured single-hand motion of 1 subject with 6 different backgrounds and lighting. Stereo and depth images are provided, but only the RGB images from the left camera are used in this work. We follow the same evaluation protocol as \cite{Zimmermann}, training on 10 and testing on the other 2 sequences. For evaluation of the 3D pose estimation, the 3D canonical pose needs to be scaled to its actual size and transformed to its global position using the ground truth root joint. Other methods scale $\textbf{w}^{can}$ by \textbf{\textit{L}} and simply translate $\textbf{w}^{can}_{mmcp}$ to $\textbf{w}^{glob}_{mmcp}$ (Cartesian alignment), which assumes that there is no rotational discrepancy since the hand is relatively close to the center of the camera. We align our canonical hand poses to global hand poses by spherical alignment. Specifically, we scale $\textbf{w}^{can}$ by \textbf{\textit{L}}, apply translation in the z-axis and rotation in the spherical coordinate system to align $\textbf{w}^{can}_{mmcp}$ with $\textbf{w}^{glob}_{mmcp}$. In Fig. \ref{fig:stereo_eval1}, we perform various experiments for self-comparisons to justify our design choices, reporting the Area Under Curve (AUC) computed using the Percentage of Correct Keypoints (PCK), where a predicted keypoint is correct if its location is within the threshold radius around the ground truth. We also report the End Point Error (EPE) in mm. For comparison with other methods, we show in Fig. \ref{fig:stereo_eval2} that we outperform most state-of-the-art methods with an AUC = 0.995. It is worth mentioning that fair comparison cannot be made since many methods utilized depth information \cite{Ge2,Cai,Iqbal}, deformable hand model (guaranteed physical plausibility) \cite{Ge2,Baek} or additional datasets \cite{Ge2,Baek,Xiang} during training while we simply trained on the left RGB images in the training set of STB with color augmentation.\\
\begin{table}[t]
  \centering
  \begin{tabular}{P{2.5cm}|c}
  \toprule
Method 	&
		\begin{tabular}{@{}c@{}}3D Hand Pose Estimation\\ 
		\begin{tabular}{P{1cm}P{3cm}}
		AUC $\uparrow$&\begin{tabular}{@{}c@{}}EPE (mm)\\\begin{tabular}{P{1.3cm}P{1.1cm}}median $\downarrow$&\ mean $\downarrow$\end{tabular}\end{tabular}
		\end{tabular}
		\end{tabular}\\
\midrule
Iqbal \etal \cite{Iqbal}&\begin{tabular}{P{1cm}P{1.4cm}P{1.2cm}}0.940&11.33&13.41\end{tabular}\\
Baek \etal \cite{Baek}&\begin{tabular}{P{1cm}P{1.4cm}P{1.2cm}}0.926&\textendash&\textendash\end{tabular}\\
Ge \etal \cite{Ge2}&\begin{tabular}{P{1cm}P{1.4cm}P{1.2cm}}0.920&\textendash&\textendash\end{tabular}\\
Cai \cite{Cai}&\begin{tabular}{P{1cm}P{1.4cm}P{1.2cm}}0.887&\textendash&\textendash\end{tabular}\\
Yang \etal \cite{Yang}&\begin{tabular}{P{1cm}P{1.4cm}P{1.2cm}}0.849&\textendash&19.95\end{tabular}\\
Spurr \etal \cite{Spurr}&\begin{tabular}{P{1cm}P{1.4cm}P{1.2cm}}0.849&\textendash&19.73\end{tabular}\\
Z \& B \cite{Zimmermann}&\begin{tabular}{P{1cm}P{1.4cm}P{1.2cm}}\textendash&18.80&\textendash\end{tabular}\\\hline
Ours&\begin{tabular}{P{1cm}P{1.4cm}P{1.2cm}}0.929&11.86&13.47\end{tabular}\\
\textbf{Ours + seg}&\begin{tabular}{P{1cm}P{1.4cm}P{1.2cm}}\textbf{0.942}&\textbf{11.14}&\textbf{12.47}\end{tabular}\\\hline
    \end{tabular}
\caption{Comparison with the state-of-the-art methods on the RHP dataset. 3D AUCs are computed over an error range from 20 to 50mm.}
\label{tab:table_rhp}
\end{table}
\indent Rendered Hand Pose Dataset (RHP) provides 41258 images for training and 2728 images for evaluation. Each rendered image contains a single character performing 1 of 39 gestures and the view is focused on one of the two hands. The training and test sets contain 31 and 8 distinct gestures respectively. We use the same setting as our method for the STB dataset. Since the pose variation in this dataset is very limited, we perform data augmentation by rotating the images with random angles. As shown in Tab. \ref{tab:table_rhp}, we achieve an AUC of 0.929 and a top AUC of 0.942 with utilization of the segmentation information. Note that \cite{Iqbal,Cai} leverage on depth data (more information than segmentation since the background has infinite depth) for training and \cite{Ge2,Baek} utilize a deformable hand model and additional datasets. Our RGB-only method outperforms other state-of-the-art methods that utilize various additional information.
\begin{figure*}[t]
  \begin{subfigure}[t]{0.29283019\linewidth}
    \includegraphics[width=\linewidth]{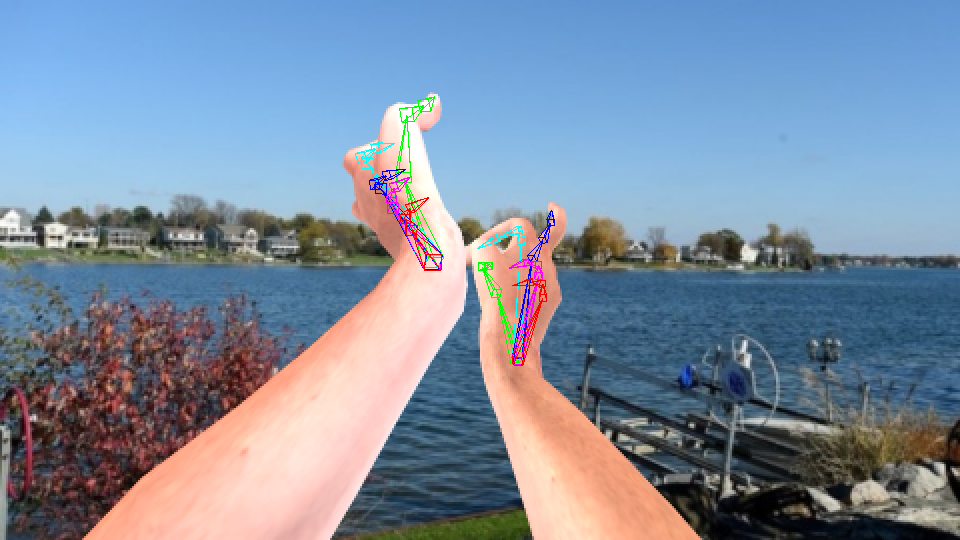}
  \end{subfigure}
  \begin{subfigure}[t]{0.29283019\linewidth}
    \includegraphics[width=\linewidth]{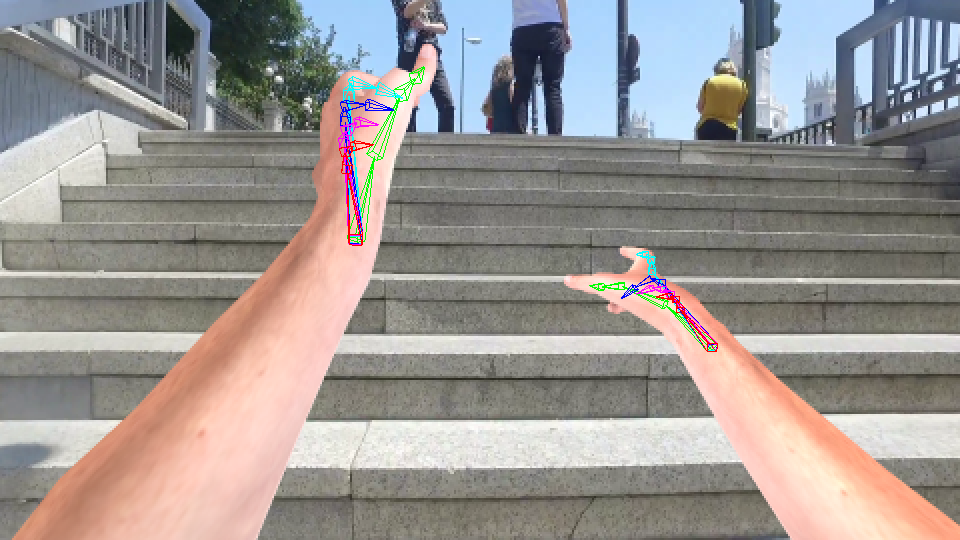}
  \end{subfigure}
  \begin{subfigure}[t]{0.21962264\linewidth}
    \includegraphics[width=\linewidth]{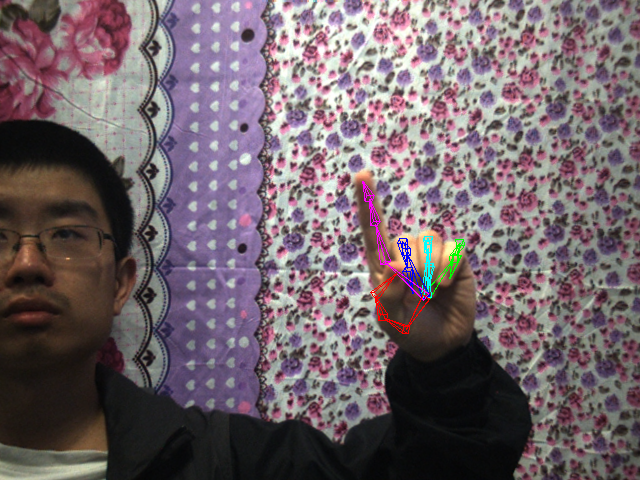}
  \end{subfigure}
  \begin{subfigure}[t]{0.16471698\linewidth}
    \includegraphics[width=\linewidth]{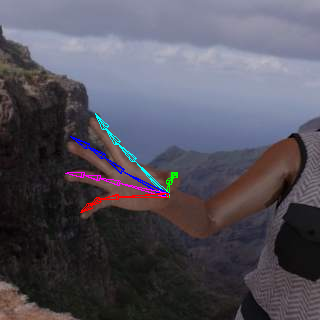}
  \end{subfigure}\\
  \begin{subfigure}[t]{0.29283019\linewidth}
    \includegraphics[width=\linewidth]{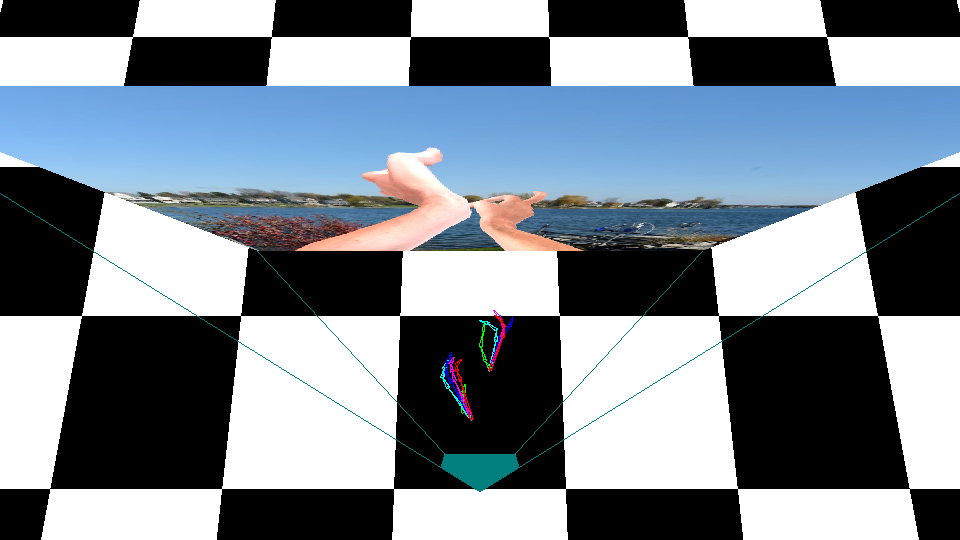}
  \end{subfigure}
  \begin{subfigure}[t]{0.29283019\linewidth}
    \includegraphics[width=\linewidth]{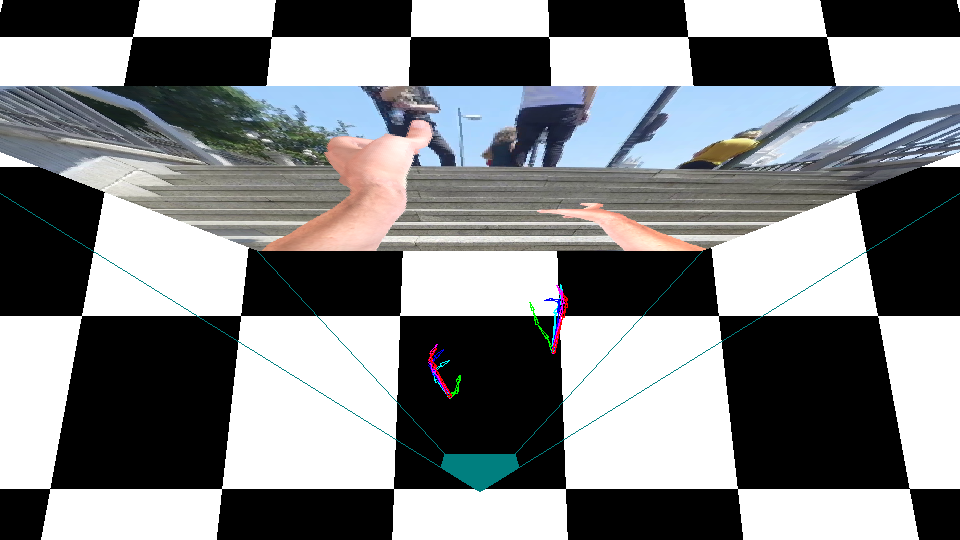}
  \end{subfigure}
  \begin{subfigure}[t]{0.21962264\linewidth}
    \includegraphics[width=\linewidth]{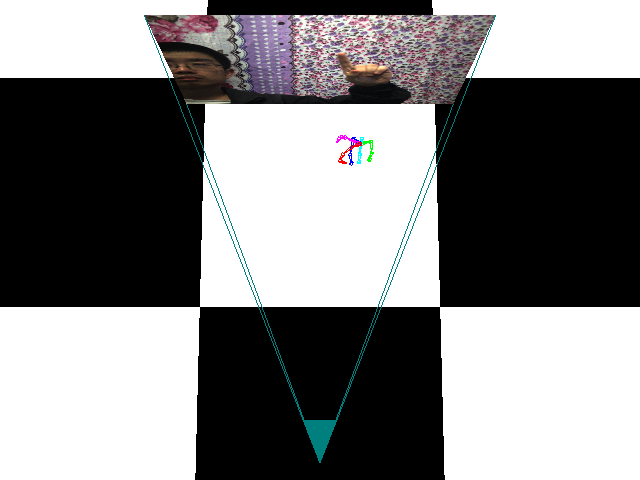}
  \end{subfigure}
  \begin{subfigure}[t]{0.16471698\linewidth}
    \includegraphics[width=\linewidth]{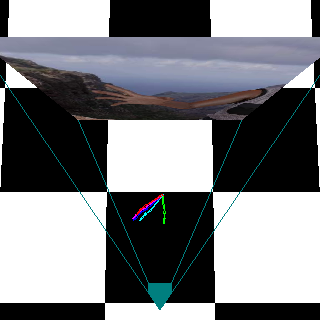}
  \end{subfigure}\\
  \begin{subfigure}[t]{0.29283019\linewidth}
    \includegraphics[width=\linewidth]{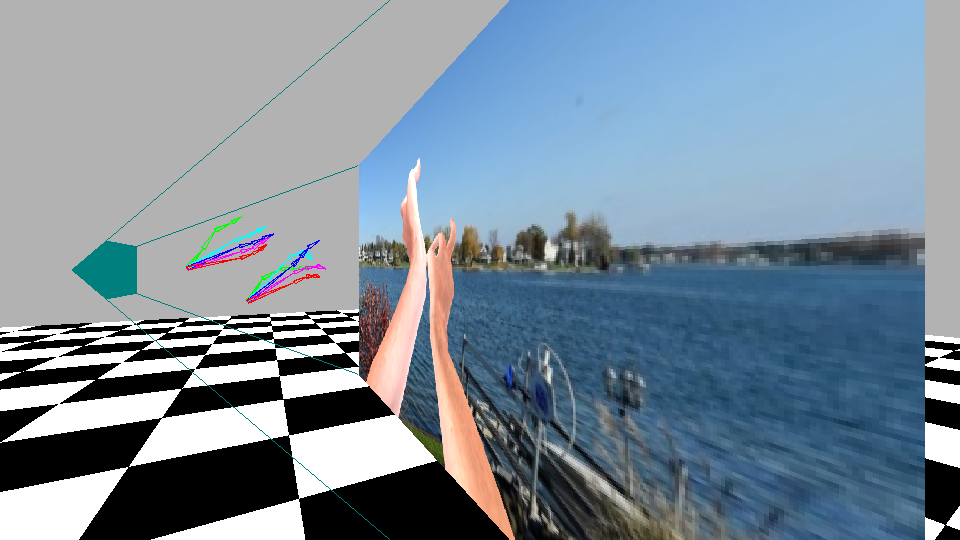}
    \caption{$Ego3D_{s}$}\label{fig:qualitative_ego_s}
  \end{subfigure}
  \begin{subfigure}[t]{0.29283019\linewidth}
    \includegraphics[width=\linewidth]{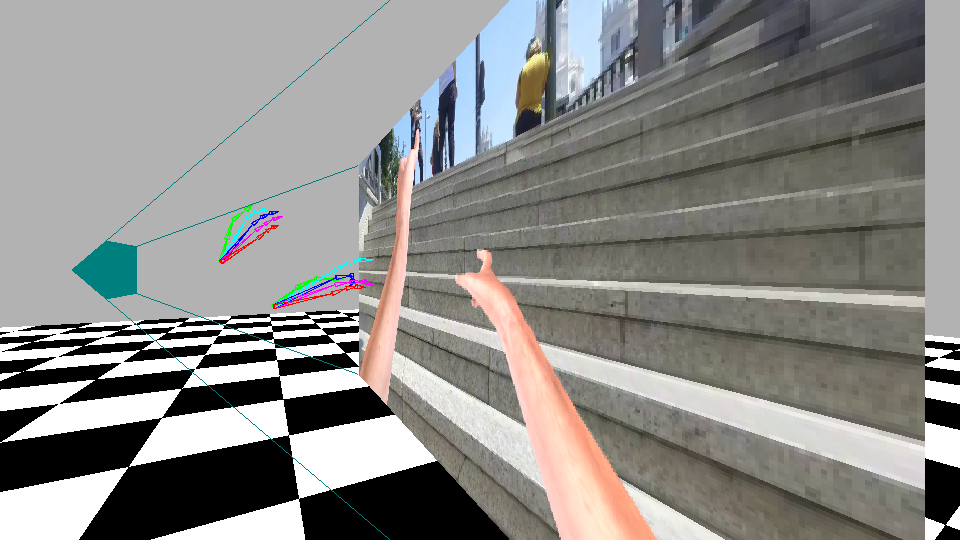}
    \caption{$Ego3D_{d}$}\label{fig:qualitative_ego_d}
  \end{subfigure}
  \begin{subfigure}[t]{0.21962264\linewidth}
    \includegraphics[width=\linewidth]{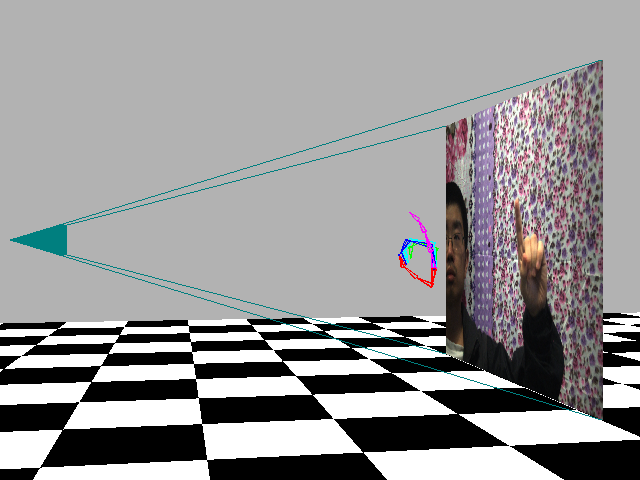}
    \caption{STB}\label{fig:qualitative_stb}
  \end{subfigure}
  \begin{subfigure}[t]{0.16471698\linewidth}
    \includegraphics[width=\linewidth]{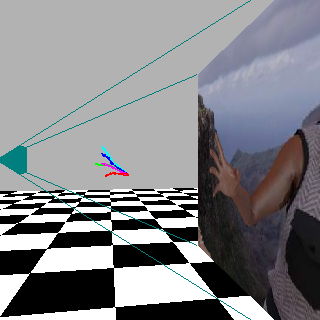}
    \caption{RHP}\label{fig:qualitative_rhp}
  \end{subfigure}
  \vspace{0.2cm}
  \caption{Qualitative results for 3D global hand pose estimation on 4 datasets. Top row visualizes the 3D global hand poses from the center camera view. Middle and bottom rows show the top and side views respectively.}
  \label{fig:qualitative_results}
\end{figure*}
\subsection{Two-hand Global Pose Estimation}\label{sec:experiments_glob}
\indent We first provide quantitative results on $Ego3D_{s}$ for all relevant subparts using the ground truth inputs for isolated analysis, then provide evaluation on the complete cascaded pipeline on both $Ego3D_{s}$ and $Ego3D_{d}$. Results for the two hands are combined by taking the average for simplicity. Extensive ablation studies are also performed.\\
\textbf{\textit{HandSegNet}}. For the segmentation of the two hands, we report mean Intersection over Union (mIoU) of 0.955 and 0.962 on $Ego3D_{s}$-test and $Ego3D_{d}$-test respectively. For hand detection, the ground truth bounding boxes are determined by the annotated 2D joint locations. We report 2 metrics for the task of hand detection: the hand detection accuracy for how well the model correctly classifies the presence of the left and right hand; the bounding box detection accuracy for how accurately the model determines the location of the left and right hands. We report a hand detection accuracy of 1.00 and bounding box detection accuracy of 0.965 and 0.982 for $Ego3D_{s}$-test and $Ego3D_{d}$-test respectively, where a positive is scored when the IoU between the ground truth and the predicted bounding boxes is greater than 0.5.\\
$\textbf{\textit{PoseNet}}_{\textbf{2D}}$. We compute 2D PCK using the ground truth and the predicted global 2D joint pixel locations. Fig. \ref{fig:ego3d_eval1} shows the 2D PCK of \textit{$PoseNet_{2D}$} on $Ego3D_{s}$-test and $Ego3D_{d}$-test. Additionally, we perform ablation studies by comparing the 2D PCK of \textit{$PoseNet_{2D}$} on various settings. We show that the hand segmentation and inserted batch normalization layers both lead to noticeable improvement.\\
$\textbf{\textit{PoseNet}}_{\textbf{3D}}$. We show the canonical 3D PCK of \textit{$PoseNet_{3D}$} on $Ego3D_{s}$-test and $Ego3D_{d}$-test in Fig. \ref{fig:ego3d_eval2}. We transform the 3D canonical hand poses to the global 3D space with spherical alignment for evaluation. We also perform ablation studies by comparing the results of training using different training losses and heatmaps. We point out that $\mathcal{L}_{bone}$ decreases the average bone length error from 3.7mm to 1.8mm despite having slightly worse AUC and EPE.\\
\textbf{\textit{3D Global Pose Estimation}}. For this new task, it is necessary that we evaluate the global pose estimation accuracy using PCK in the spherical coordinate system for more intuitive results. Specifically, the spherical PCK evaluates directional accuracy and distance accuracy of the root joint (mMCP). We claim that the spherical PCK on the root joint and PCK for the 3D canonical pose estimation together produce comprehensive evaluation results for the task of 3D global hand pose estimation. We skip the spherical PCK plot for isolated study since our 3D global pose estimation algorithm perfectly reconstructs the global 3D poses given the ground truth $\textbf{w}^{can}$ in 3D and $\textbf{p}$ in 2D.\\
\indent The task for the complete cascaded pipeline is particularly difficult not only due to each module being dependent on the accuracy of the prior estimation, but also the fact that any error on the 2D or 3D canonical pose estimation can directly impact the global projection and decrease the final accuracy. Fig. \ref{fig:ego3d_eval3} shows the spherical PCK of our complete pipeline on $Ego3D_{s}$-test and $Ego3D_{d}$-test. Note that $\mathcal{L}_{proj}$ improved the overall spherical PCK for 3D global pose estimation despite leading to slightly worse performance in 3D canonical pose estimation.\\
\indent We demonstrate that global hand pose estimation through monocular RGB input is achievable and we show promising results. For $Ego3D_{s}$ and $Ego3D_{d}$, our method achieves a directional and distance accuracy of 0.90 approximately at an angular threshold of 3 degrees and a radius threshold of 7 cm respectively. Note that hand poses with differences of 7 cm in distance with respect to the camera origin show little difference visually in 2D images but there is definitely room for improvement. \\
\indent Our global estimation algorithm also generalizes for other datasets. We report the spherical PCK on STB in Fig. \ref{fig:stereo_eval3} for global pose estimation without accessing the ground truth location of the root joint. Our results on STB indicates that our method is capable of accurate global pose estimation on real-world data as well if sufficient training data is available. For RHP, we report an AUC of 0.960, 0.958 and 0.690 for the spherical PCK of $\theta$, $\phi$ and radius respectively. This dataset is challenging for accurate distance estimation due to its low image resolution. We show qualitative results for 3D global hand pose estimation on 4 datasets in Fig. \ref{fig:qualitative_results}. We provide additional qualitative results and preliminary evaluation on real-world data in the supplementary material.
\section{Conclusion}
\indent In this work we present the first method that estimates the 3D global poses for both hands given only a single RGB image. We contribute a large-scale synthetic egocentric hand pose dataset for training and evaluation of the networks. We show that our approach outperforms methods that utilize additional information for single-hand 3D canonical hand pose estimation and further achieves promising results for two-hand 3D global hand pose estimation. Evaluation on real-world data remains as necessary future work, which requires sufficiently annotated training data in the real-world domain.

\nocite{Liu}
{\small
\bibliographystyle{ieee_fullname}
\bibliography{egbib}
}

\clearpage
\onecolumn
\title{\vspace{-2.0cm}Supplementary Document:\\
Two-hand Global 3D Pose Estimation using Monocular RGB
}

\author{}
\date{}
\maketitle
\vspace{-3.0cm}
\setcounter{section}{0}
\section{Training Details}
\indent To obtain the experimental results for all targeting datasets, we use the same training schedule for \textit{HandSegNet}, \textit{$PoseNet_{2D}$} and \textit{$PoseNet_{3D}$}. Specifically, we use the Adam optimizer with an initial learning rate of 0.001, $\beta_{1}$ = 0.9 and $\beta_{2}$ = 0.999. We use cross entropy (CE) loss for the segmentation loss of \textit{HandSegNet} and mean squared error (MSE) loss for the training of all other parts of the networks. We set the batch size to 4 and  trained \textit{HandSegNet} for 30,000 iterations. \textit{$PoseNet_{2D}$} and \textit{$PoseNet_{3D}$} are trained for 15,000 iterations since each iteration consists of training of two separate hands. The learning rates decrease with a rate of 0.5 every 5,000 iterations.\\
\indent For results obtained without segmentation and batch normalization layers using \textit{$PoseNet_{2D}$}, we used standard stochastic gradient descent and an initial learning rate of 0.000001 for better convergence. We also set the weight decay to 0.0005 for $PoseNet_{3D}$.
\section{Additional Qualitative Results}
\indent For evaluation on real-world data, we manually annotated a small dataset to train the networks and provide preliminary qualitative results in Fig. \ref{fig:qualitative_results1}. We evaluate on simple poses due to the limited variety in our annotated data. Our preliminary results indicate that our method is capable of evaluation on real-world data when sufficient training data in the real-world domain becomes available in the future. Note that evaluation on the RGB data in the real-world domain is extremely challenging due to factors such as the vast color space, different skin color/texture, complex background noise, motion blur, lighting, shadow features, etc. As a result, evaluation on videos in the wild is beyond the scope of this paper and will be addressed in our future works. It is worth mentioning that since it is currently infeasible to quantitatively evaluate on the task of RGB-based two-hand global 3D pose estimation using real-world data due to the lack of ground truth data, Ego3DHands serves as a necessary benchmark dataset for this new task. We also provide additional qualitative results for the 4 target datasets in Fig. \ref{fig:qualitative_results2}.
\setcounter{figure}{0}
\begin{figure*}[h]
  \begin{subfigure}[t]{0.32\linewidth}
    \includegraphics[width=\linewidth]{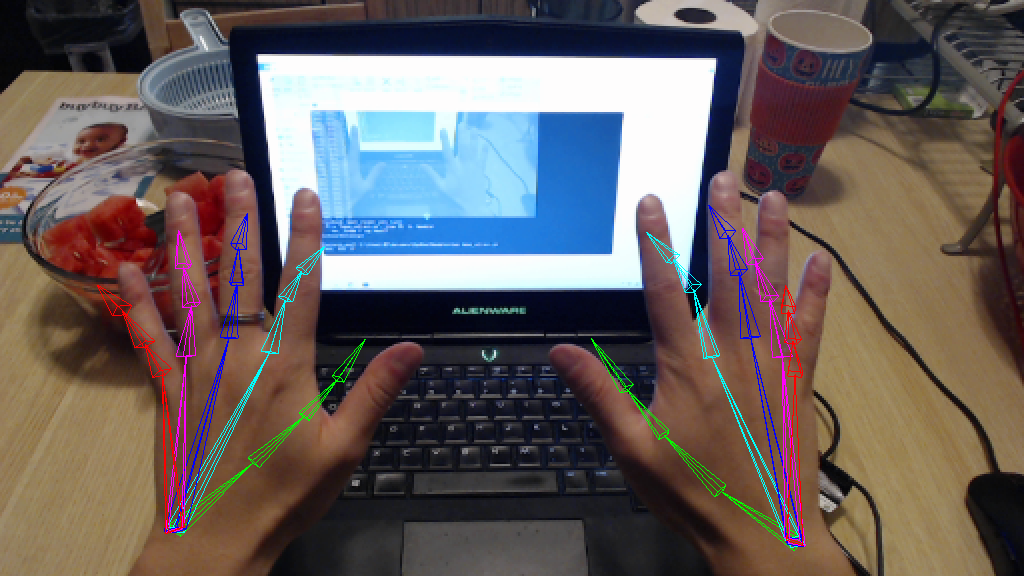}
  \end{subfigure}
  \begin{subfigure}[t]{0.32\linewidth}
    \includegraphics[width=\linewidth]{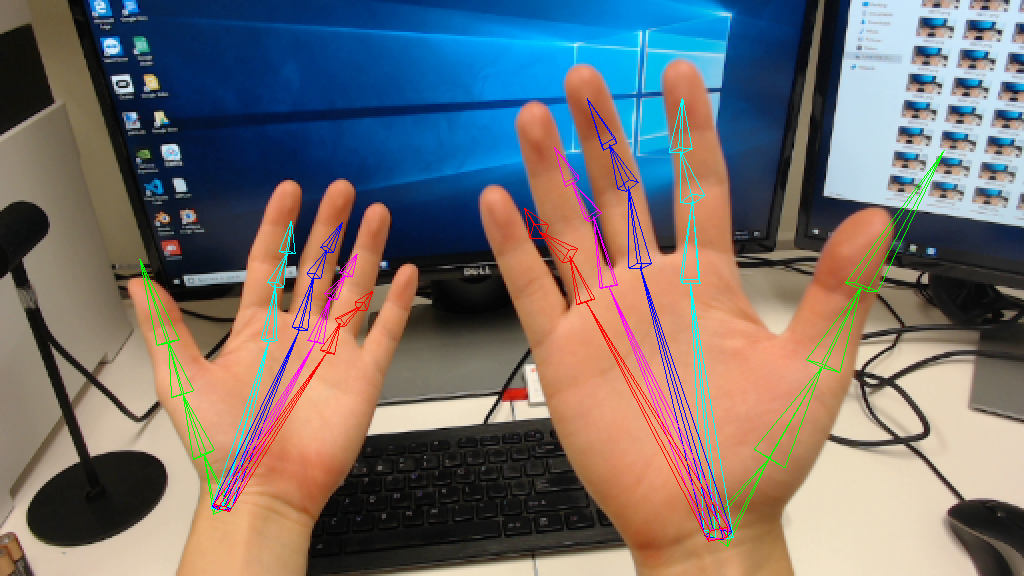}
  \end{subfigure}
  \begin{subfigure}[t]{0.32\linewidth}
    \includegraphics[width=\linewidth]{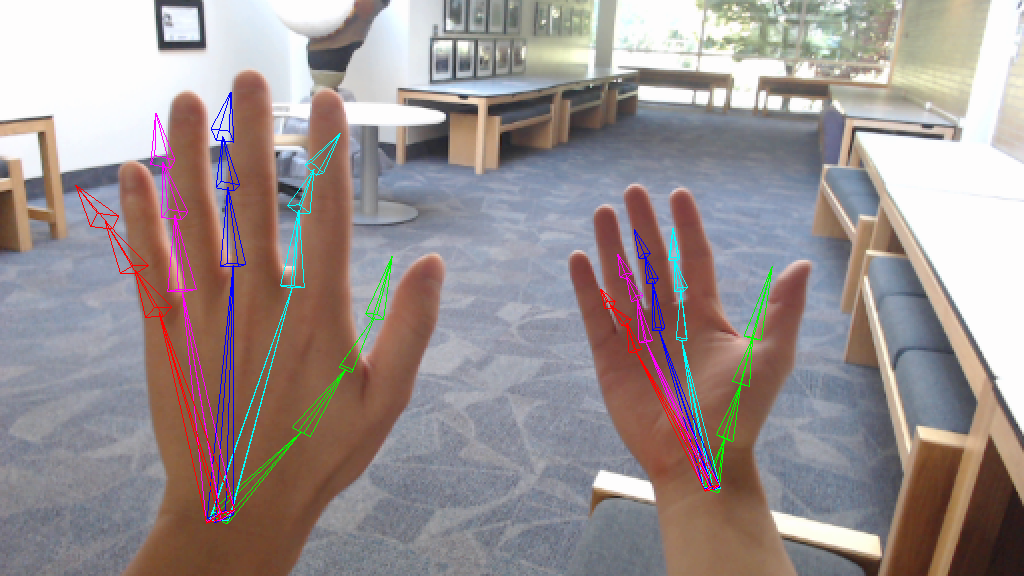}
  \end{subfigure}\\
  \begin{subfigure}[t]{0.32\linewidth}
    \includegraphics[width=\linewidth]{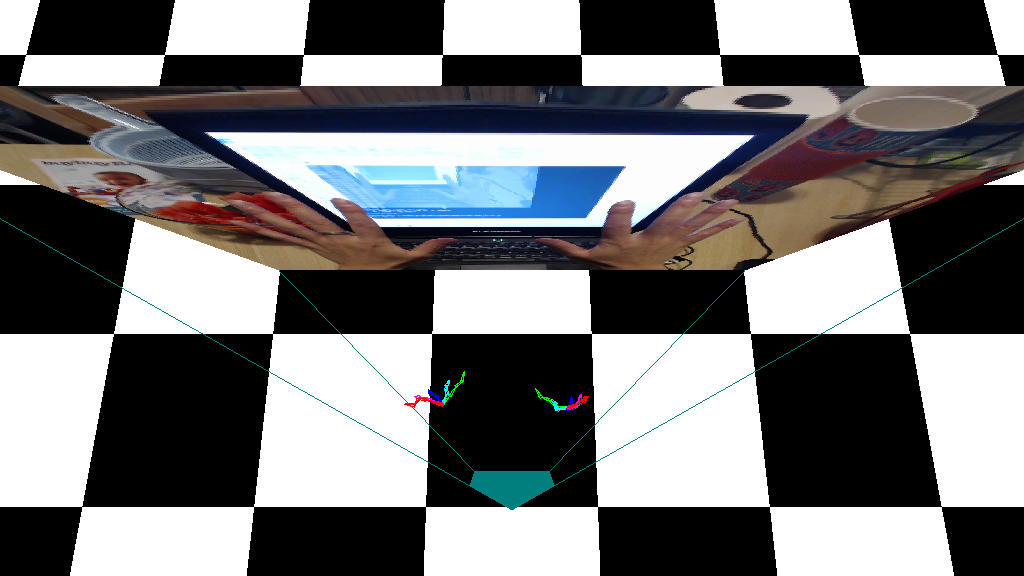}
  \end{subfigure}
  \begin{subfigure}[t]{0.32\linewidth}
    \includegraphics[width=\linewidth]{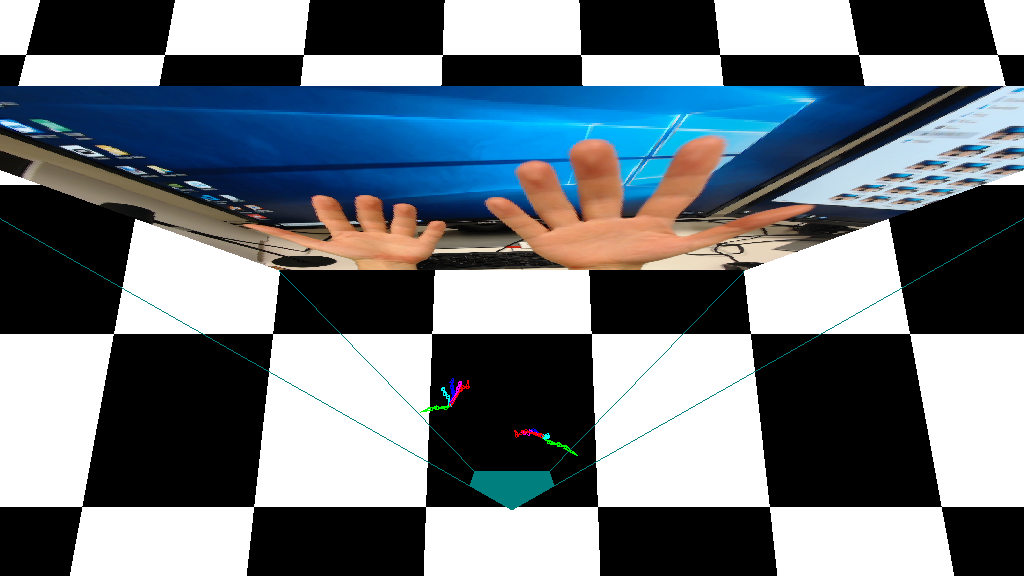}
  \end{subfigure}
  \begin{subfigure}[t]{0.32\linewidth}
    \includegraphics[width=\linewidth]{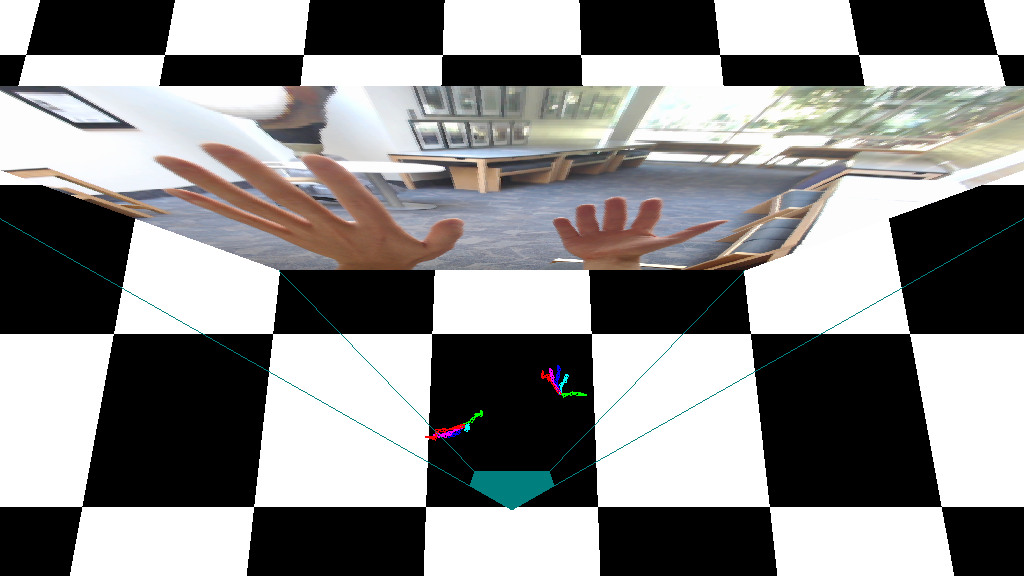}
  \end{subfigure}\\
  \begin{subfigure}[t]{0.32\linewidth}
    \includegraphics[width=\linewidth]{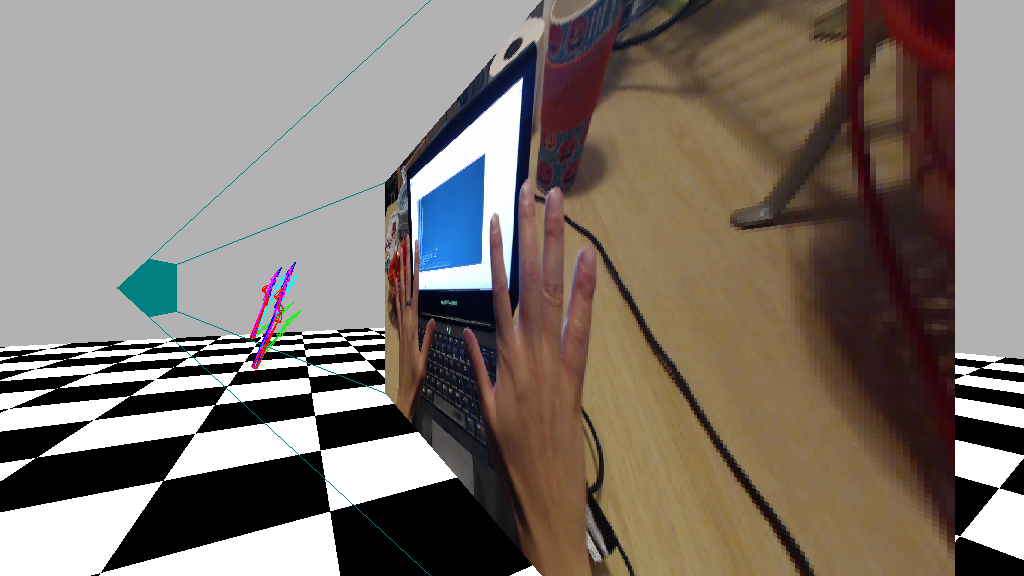}
  \end{subfigure}
  \begin{subfigure}[t]{0.32\linewidth}
    \includegraphics[width=\linewidth]{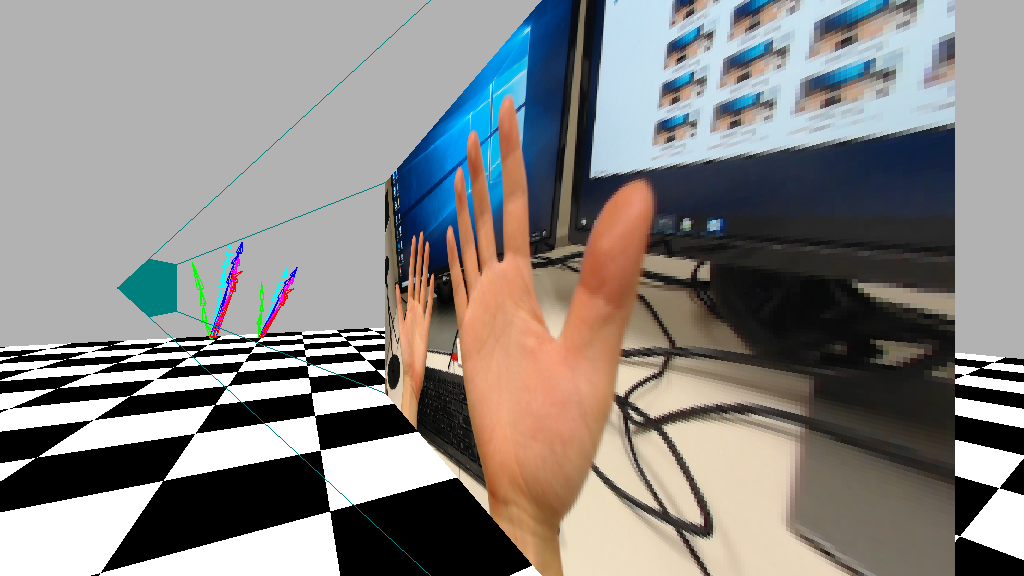}
  \end{subfigure}
  \begin{subfigure}[t]{0.32\linewidth}
    \includegraphics[width=\linewidth]{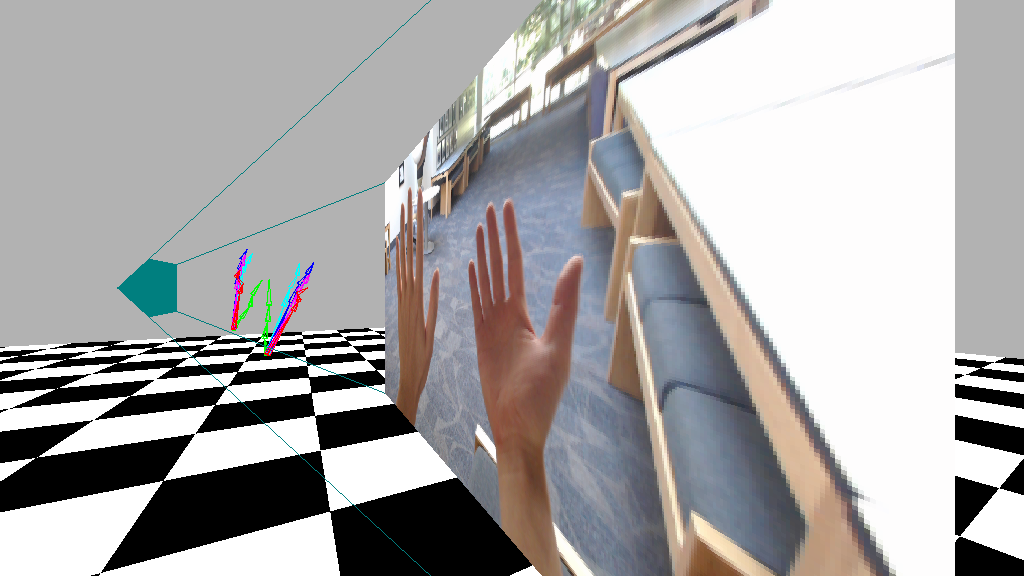}
  \end{subfigure}
  \vspace{0.2cm}
  \caption{Preliminary qualitative results on real-world data. We collected sample test sequences using 3 different background environments. Top row visualizes the 3D global hand poses from the center camera view. Middle and bottom rows show the top and side views respectively.}
  \label{fig:qualitative_results1}
\end{figure*}
\begin{figure*}[h]
  \begin{subfigure}[t]{0.29283019\linewidth}
    \includegraphics[width=\linewidth]{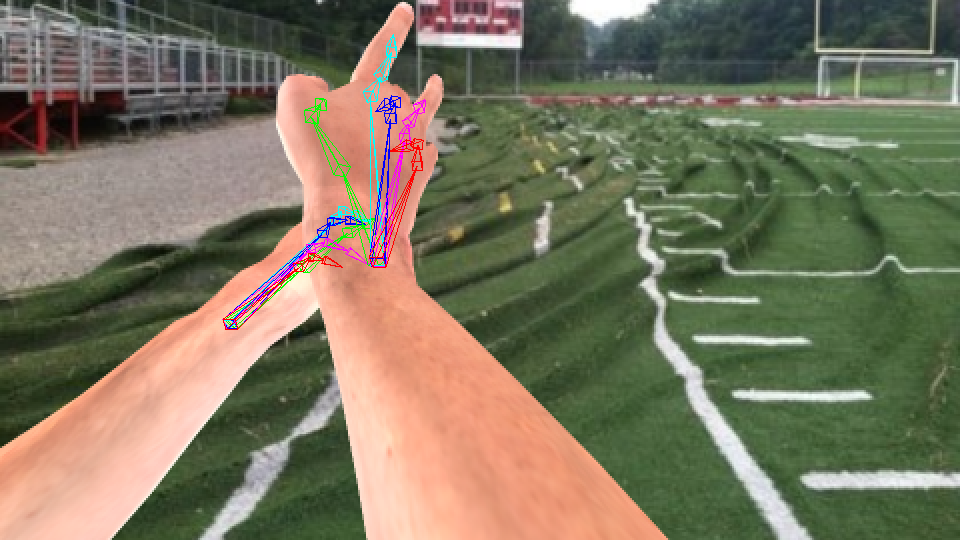}
  \end{subfigure}
  \begin{subfigure}[t]{0.29283019\linewidth}
    \includegraphics[width=\linewidth]{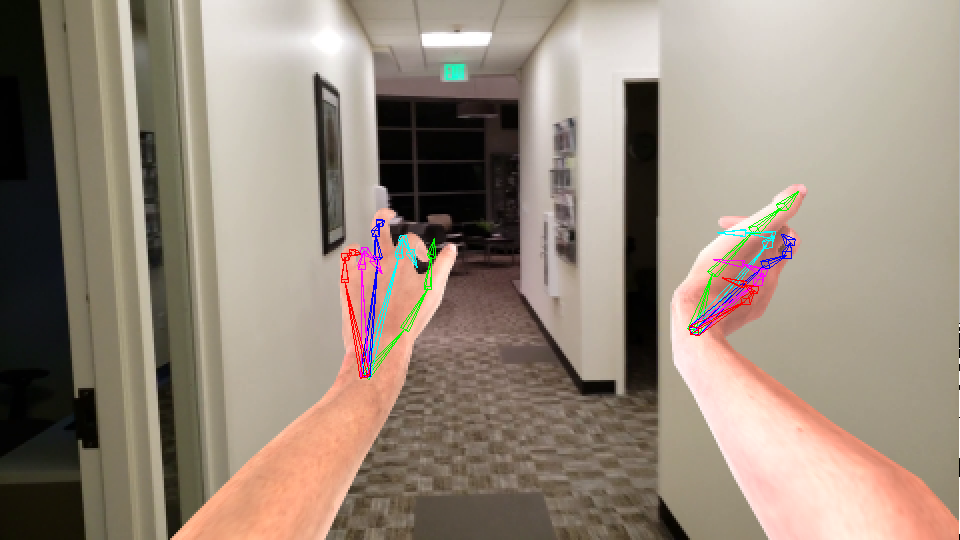}
  \end{subfigure}
  \begin{subfigure}[t]{0.21962264\linewidth}
    \includegraphics[width=\linewidth]{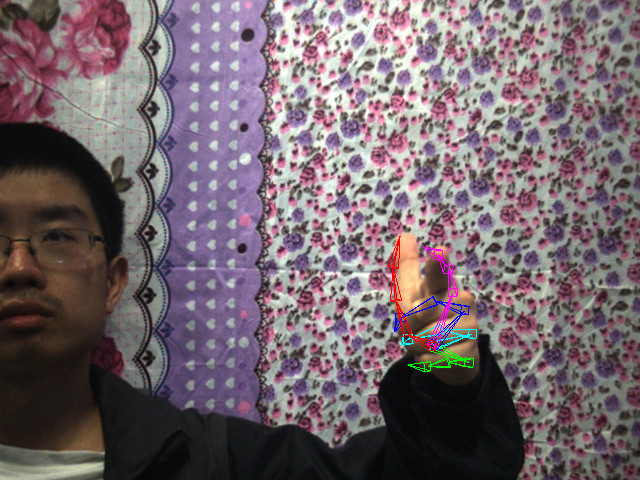}
  \end{subfigure}
  \begin{subfigure}[t]{0.16471698\linewidth}
    \includegraphics[width=\linewidth]{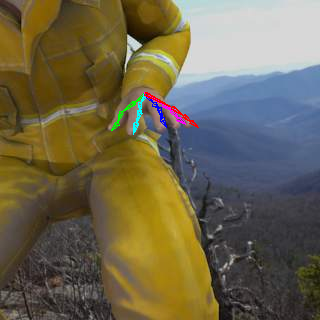}
  \end{subfigure}\\
  \begin{subfigure}[t]{0.29283019\linewidth}
    \includegraphics[width=\linewidth]{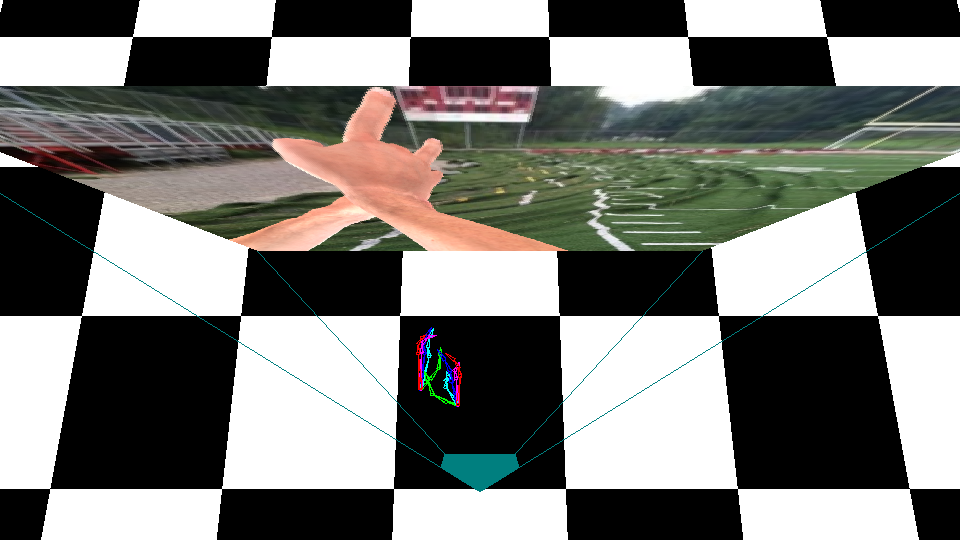}
  \end{subfigure}
  \begin{subfigure}[t]{0.29283019\linewidth}
    \includegraphics[width=\linewidth]{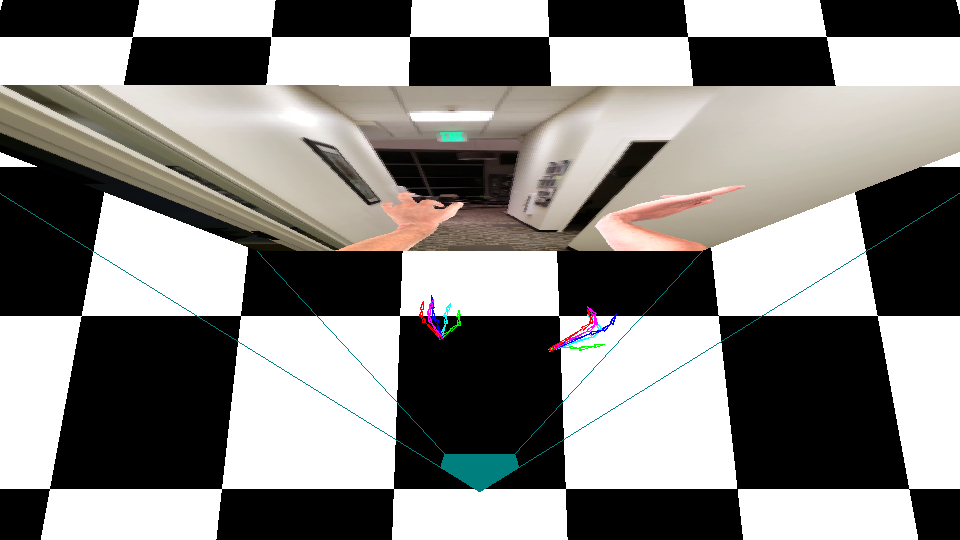}
  \end{subfigure}
  \begin{subfigure}[t]{0.21962264\linewidth}
    \includegraphics[width=\linewidth]{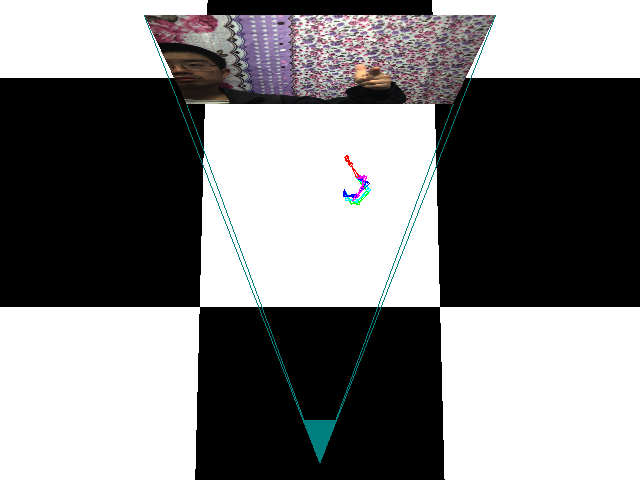}
  \end{subfigure}
  \begin{subfigure}[t]{0.16471698\linewidth}
    \includegraphics[width=\linewidth]{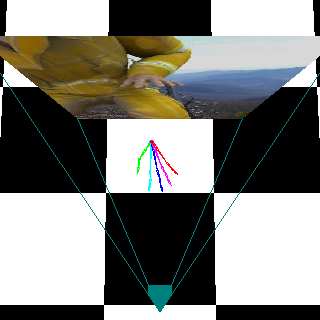}
  \end{subfigure}\\
  \begin{subfigure}[t]{0.29283019\linewidth}
    \includegraphics[width=\linewidth]{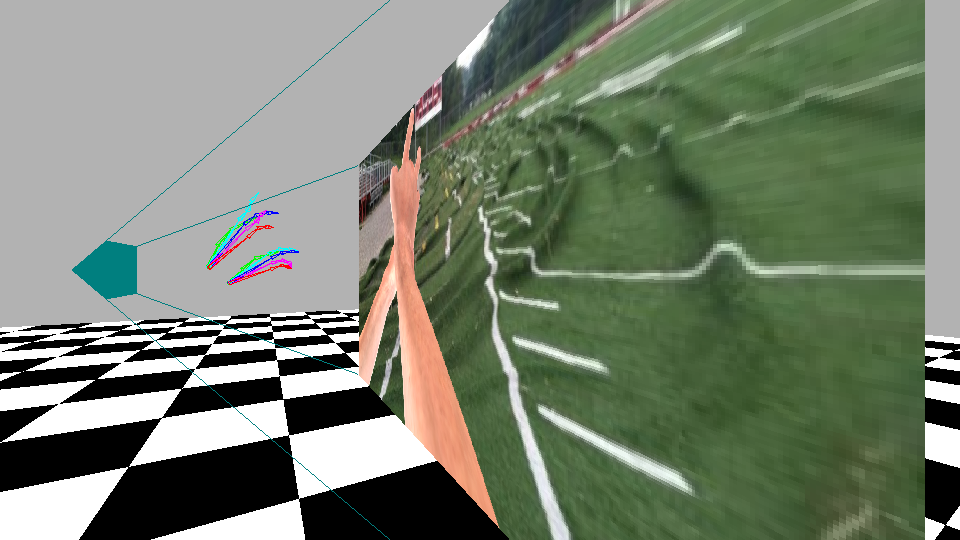}
    \caption{$Ego3D_{s} (img_{1})$}\label{fig:qualitative_ego_s}
  \end{subfigure}
  \begin{subfigure}[t]{0.29283019\linewidth}
    \includegraphics[width=\linewidth]{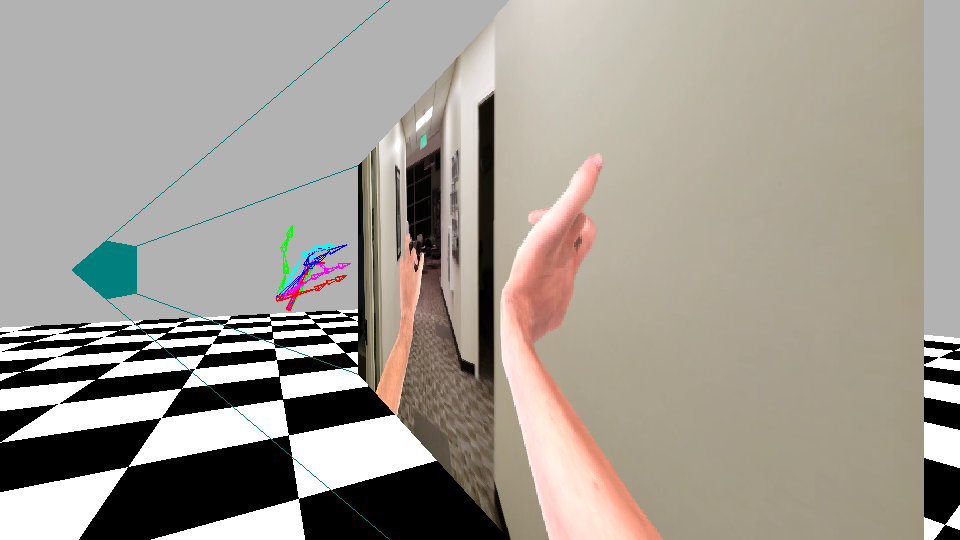}
    \caption{$Ego3D_{d} (img_{1})$}\label{fig:qualitative_ego_d}
  \end{subfigure}
  \begin{subfigure}[t]{0.21962264\linewidth}
    \includegraphics[width=\linewidth]{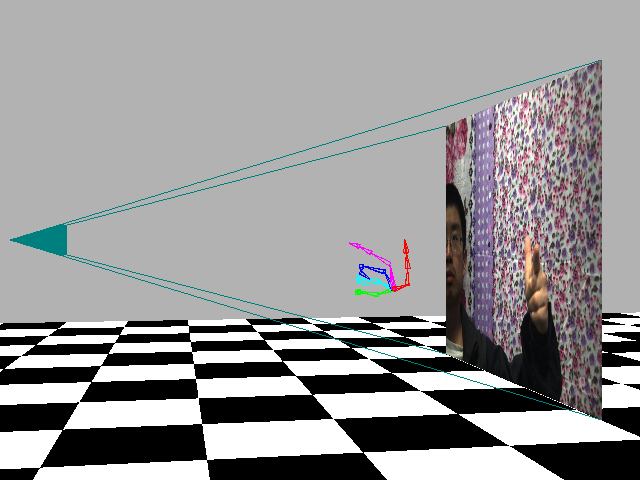}
    \caption{STB $(img_{1})$}\label{fig:qualitative_stb}
  \end{subfigure}
  \begin{subfigure}[t]{0.16471698\linewidth}
    \includegraphics[width=\linewidth]{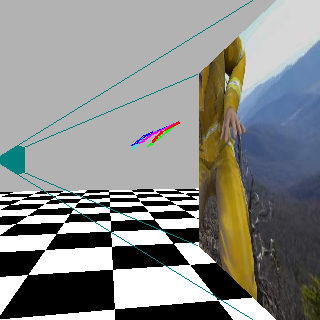}
    \caption{RHP $(img_{1})$}\label{fig:qualitative_rhp}
  \end{subfigure}
  \vspace{0.2cm}
  \vspace{-0.4cm}
\end{figure*}
\appendix
\renewcommand\thefigure{\thesection\arabic{figure}}
\setcounter{figure}{1}
\begin{figure*}[h]
  \begin{subfigure}[t]{0.29283019\linewidth}
    \includegraphics[width=\linewidth]{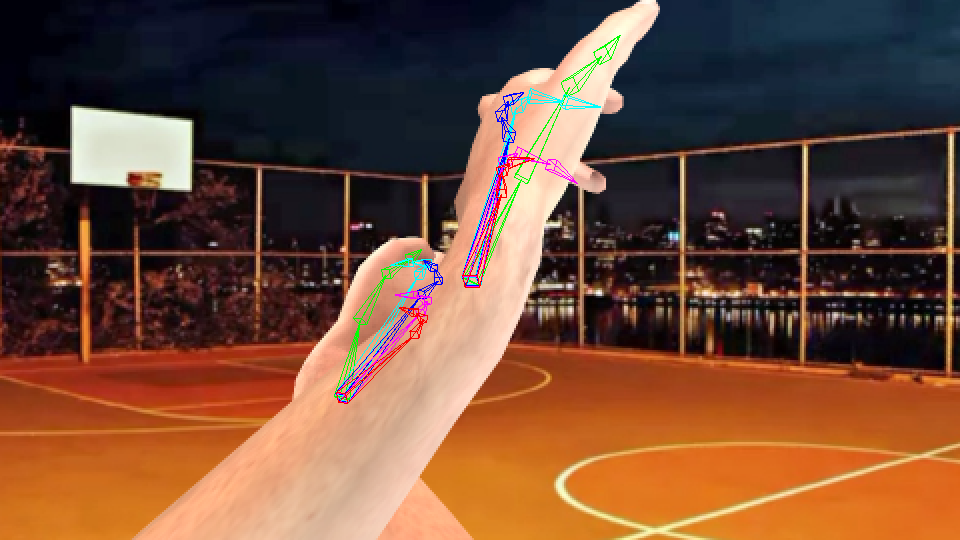}
  \end{subfigure}
  \begin{subfigure}[t]{0.29283019\linewidth}
    \includegraphics[width=\linewidth]{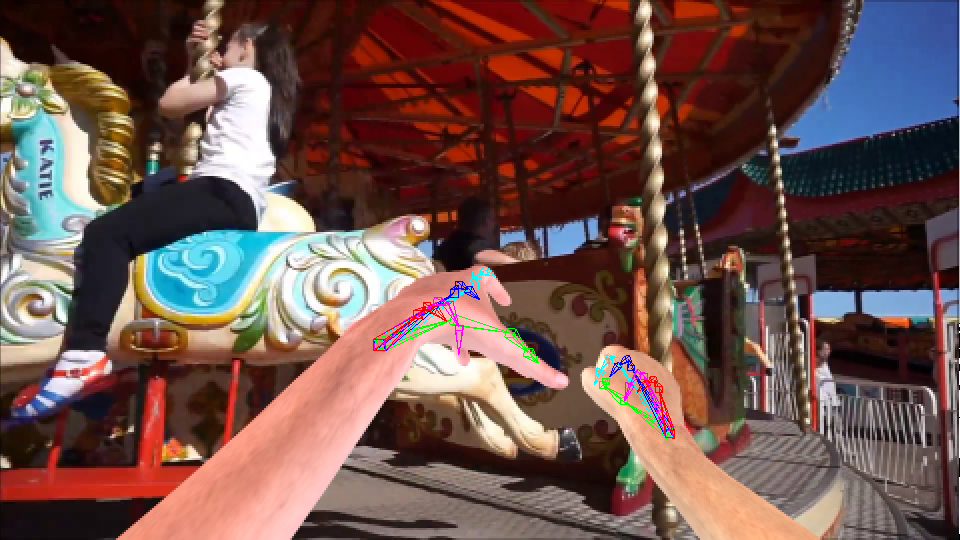}
  \end{subfigure}
  \begin{subfigure}[t]{0.21962264\linewidth}
    \includegraphics[width=\linewidth]{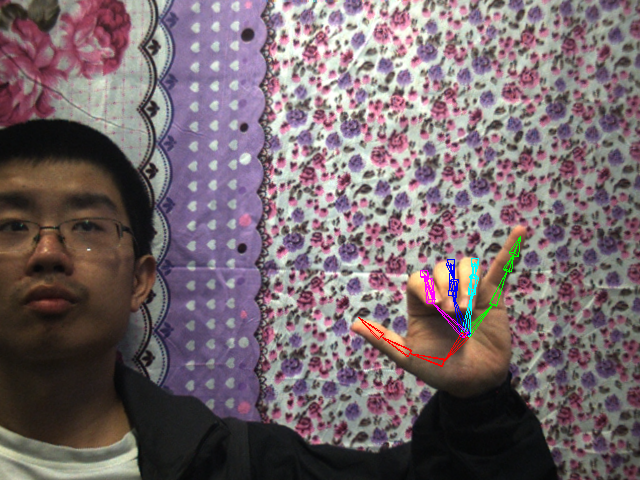}
  \end{subfigure}
  \begin{subfigure}[t]{0.16471698\linewidth}
    \includegraphics[width=\linewidth]{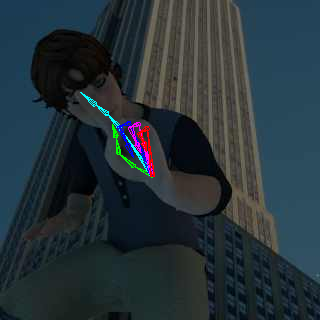}
  \end{subfigure}\\
  \begin{subfigure}[t]{0.29283019\linewidth}
    \includegraphics[width=\linewidth]{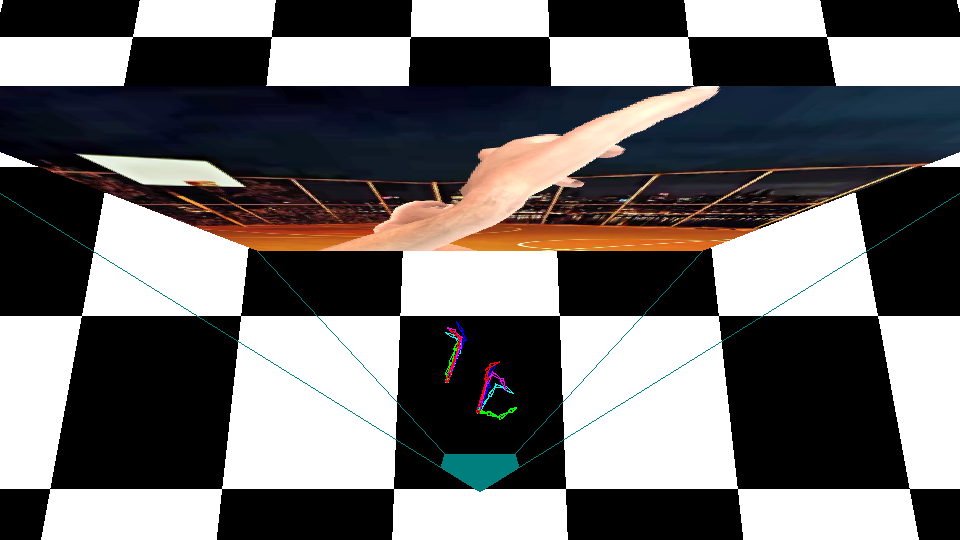}
  \end{subfigure}
  \begin{subfigure}[t]{0.29283019\linewidth}
    \includegraphics[width=\linewidth]{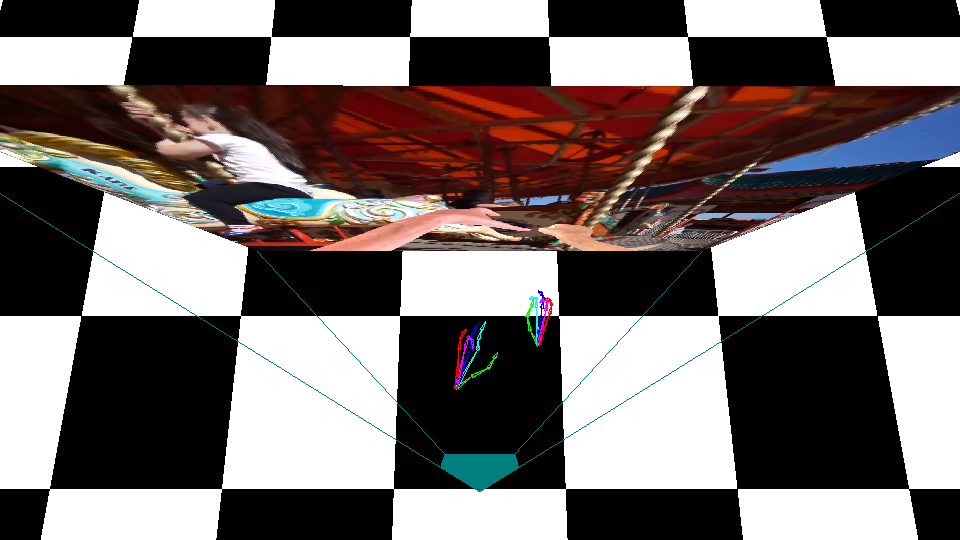}
  \end{subfigure}
  \begin{subfigure}[t]{0.21962264\linewidth}
    \includegraphics[width=\linewidth]{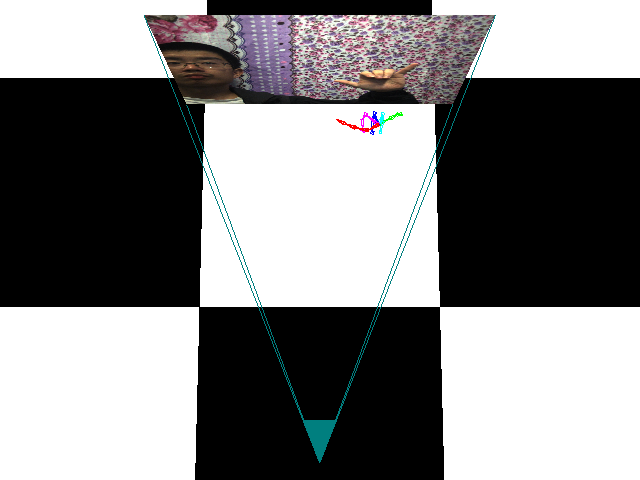}
  \end{subfigure}
  \begin{subfigure}[t]{0.16471698\linewidth}
    \includegraphics[width=\linewidth]{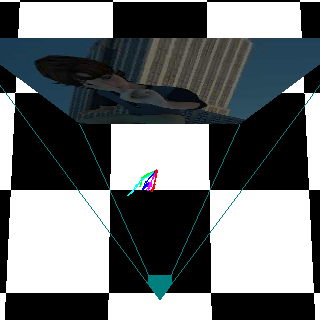}
  \end{subfigure}\\
  \begin{subfigure}[t]{0.29283019\linewidth}
    \includegraphics[width=\linewidth]{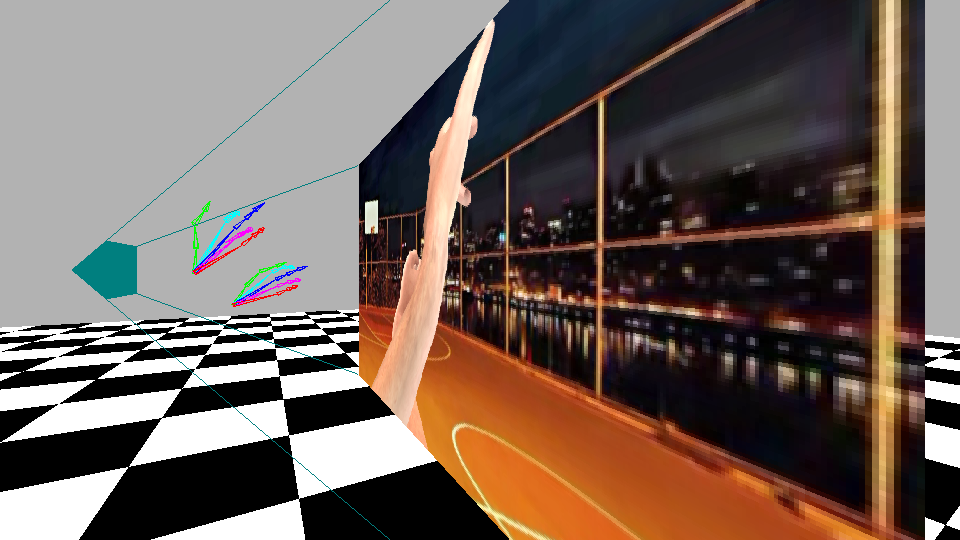}
    \caption{$Ego3D_{s} (img_{2})$}\label{fig:qualitative_ego_s}
  \end{subfigure}
  \begin{subfigure}[t]{0.29283019\linewidth}
    \includegraphics[width=\linewidth]{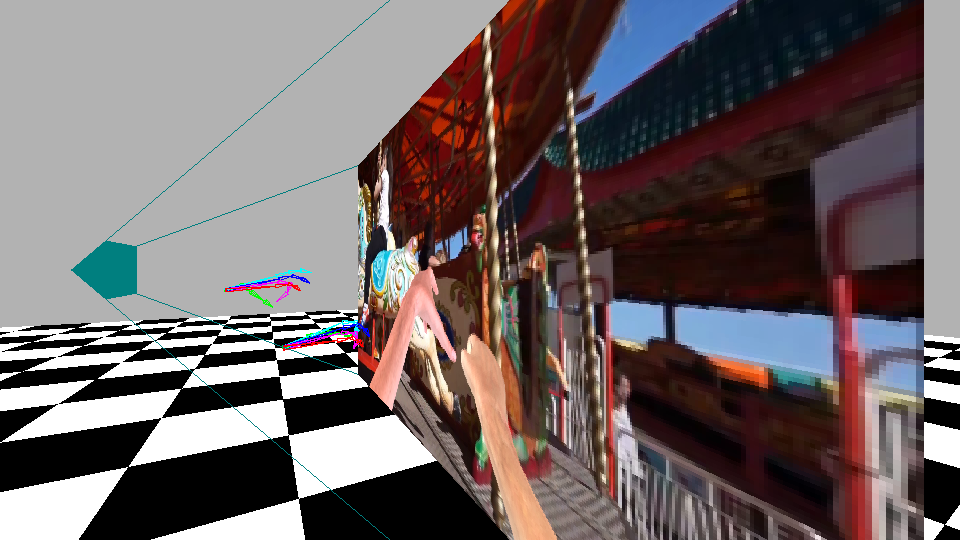}
    \caption{$Ego3D_{d} (img_{2})$}\label{fig:qualitative_ego_d}
  \end{subfigure}
  \begin{subfigure}[t]{0.21962264\linewidth}
    \includegraphics[width=\linewidth]{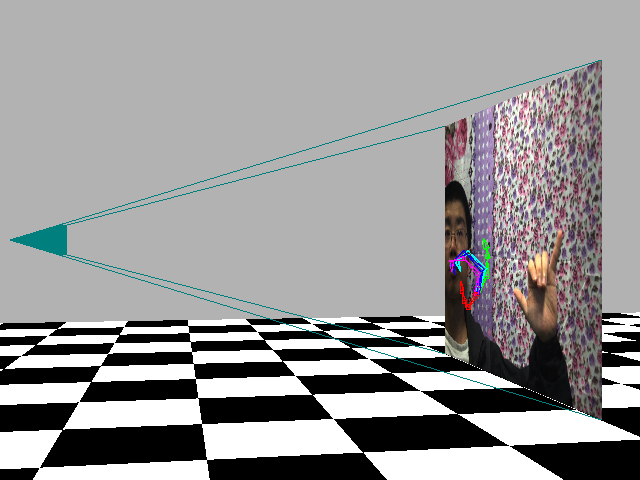}
    \caption{STB $(img_{2})$}\label{fig:qualitative_stb}
  \end{subfigure}
  \begin{subfigure}[t]{0.16471698\linewidth}
    \includegraphics[width=\linewidth]{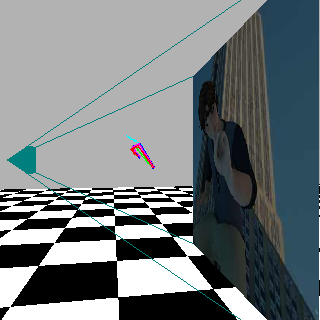}
    \caption{RHP $(img_{2})$}\label{fig:qualitative_rhp}
  \end{subfigure}
  \vspace{0.4cm}
  \caption{Additional qualitative results for 3D global two-hand pose estimation on $Ego3D_{s}$, $Ego3D_{d}$, STB and RHP. }
  \label{fig:qualitative_results2}
\end{figure*}

\end{document}